\documentclass[10pt,journal,compsoc]{IEEEtran}
\ifCLASSOPTIONcompsoc
  \usepackage[nocompress]{cite}
\else
  \usepackage{cite}
\fi
\ifCLASSINFOpdf
  \usepackage[pdftex]{graphicx}
\else
\fi
\usepackage{amsmath}
\usepackage{array}
\usepackage{url}
\usepackage[utf8]{inputenc}
\usepackage{mathptmx}
\usepackage{amsmath}
\usepackage{amssymb}
\usepackage{url}

\usepackage{subfigure}
\usepackage{algpseudocode}
\usepackage{algorithm}
\usepackage{algorithmicx}
\usepackage{xfrac}

\usepackage{xcolor}

\usepackage{graphicx}
\usepackage{grffile}

\usepackage{tikz}
\definecolor{carminered}{rgb}{1.0, 0.0, 0.22}
\newcommand*\circledr[1]{\tikz[baseline=(char.base)]{
    \node[shape=circle, fill=carminered, draw=carminered, text=white, inner sep=0.85pt] (char) {\sffamily\footnotesize\textbf{#1}};}}

\newcommand{\squishlist}{
	\begin{list}{$\bullet$}
		{ \setlength{\itemsep}{0pt}      \setlength{\parsep}{3pt}
			\setlength{\topsep}{3pt}       \setlength{\partopsep}{0pt}
			\setlength{\leftmargin}{1.5em} \setlength{\labelwidth}{1em}
			\setlength{\labelsep}{0.5em} } }
	
	\newcommand{\squishlisttwo}{
		\begin{list}{$\bullet$}
			{ \setlength{\itemsep}{0pt}    \setlength{\parsep}{0pt}
				\setlength{\topsep}{0pt}     \setlength{\partopsep}{0pt}
				\setlength{\leftmargin}{2em} \setlength{\labelwidth}{1.5em}
				\setlength{\labelsep}{0.5em} } }
		
		\newcommand{\squishend}{
		\end{list}  }

\usepackage{caption}

\begin{document}
%
% paper title
% Titles are generally capitalized except for words such as a, an, and, as,
% at, but, by, for, in, nor, of, on, or, the, to and up, which are usually
% not capitalized unless they are the first or last word of the title.
% Linebreaks \\ can be used within to get better formatting as desired.
% Do not put math or special symbols in the title.
\title{A Grid-based Method for Removing Overlaps of Dimensionality Reduction Scatterplot Layouts}
%
%
% author names and IEEE memberships
% note positions of commas and nonbreaking spaces ( ~ ) LaTeX will not break
% a structure at a ~ so this keeps an author's name from being broken across
% two lines.
% use \thanks{} to gain access to the first footnote area
% a separate \thanks must be used for each paragraph as LaTeX2e's \thanks
% was not built to handle multiple paragraphs
%
%
%\IEEEcompsocitemizethanks is a special \thanks that produces the bulleted
% lists the Computer Society journals use for "first footnote" author
% affiliations. Use \IEEEcompsocthanksitem which works much like \item
% for each affiliation group. When not in compsoc mode,
% \IEEEcompsocitemizethanks becomes like \thanks and
% \IEEEcompsocthanksitem becomes a line break with idention. This
% facilitates dual compilation, although admittedly the differences in the
% desired content of \author between the different types of papers makes a
% one-size-fits-all approach a daunting prospect. For instance, compsoc 
% journal papers have the author affiliations above the "Manuscript
% received ..."  text while in non-compsoc journals this is reversed. Sigh.

\author{Gladys M. Hilasaca, Wilson E. Marcílio-Jr, Danilo M. Eler, Rafael M. Martins, and Fernando V. Paulovich,~\IEEEmembership{Member,~IEEE}% <-this % stops a space
\IEEEcompsocitemizethanks{
\IEEEcompsocthanksitem G.M. Hilasaca is with the University of São Paulo, Brazil.  \protect\\
% note need leading \protect in front of \\ to get a newline within \thanks as
% \\ is fragile and will error, could use \hfil\break instead.
E-mail: marlenyh@icmc.usp.br
\IEEEcompsocthanksitem W.E. Marcílio-Jr and D.M. Eler are with the São Paulo State University, Brazil. \protect\\
E-mail: wilson.marcilio@unesp.br, danilo.eler@unesp.br
\IEEEcompsocthanksitem  R. M. Martins is with Linnaeus University, Sweden. \protect\\
E-mail: rafael.martins@lnu.se
\IEEEcompsocthanksitem F. V. Paulovich is with Eindhoven University of Technology (TU/e), the Netherlands.
\protect\\
E-mail: f.paulovich@tue.nl}% <-this % stops an unwanted space
\thanks{Manuscript received September 14, 2022; revised August 13, 2023.}
}
% \\ \bigskip \insertfig}
% \thanks{Manuscript received April 19, 2005; revised August 26, 2015.}}

% note the % following the last \IEEEmembership and also \thanks - 
% these prevent an unwanted space from occurring between the last author name
% and the end of the author line. i.e., if you had this:
% 
% \author{....lastname \thanks{...} \thanks{...} }
%                     ^------------^------------^----Do not want these spaces!
%
% a space would be appended to the last name and could cause every name on that
% line to be shifted left slightly. This is one of those "LaTeX things". For
% instance, "\textbf{A} \textbf{B}" will typeset as "A B" not "AB". To get
% "AB" then you have to do: "\textbf{A}\textbf{B}"
% \thanks is no different in this regard, so shield the last } of each \thanks
% that ends a line with a % and do not let a space in before the next \thanks.
% Spaces after \IEEEmembership other than the last one are OK (and needed) as
% you are supposed to have spaces between the names. For what it is worth,
% this is a minor point as most people would not even notice if the said evil
% space somehow managed to creep in.

% The paper headers
\markboth{IEEE Transactions on Visualization and Computer Graphics,~Vol.~00, No.~0, August~2023}%
{Mamani \MakeLowercase{\textit{et al.}}: A Grid-based Method for Removing Overlaps of Dimensionality Reduction Scatterplot Layouts}

% The only time the second header will appear is for the odd numbered pages
% after the title page when using the twoside option.
% 
% *** Note that you probably will NOT want to include the author's ***
% *** name in the headers of peer review papers.                   ***
% You can use \ifCLASSOPTIONpeerreview for conditional compilation here if
% you desire.

% The publisher's ID mark at the bottom of the page is less important with
% Computer Society journal papers as those publications place the marks
% outside of the main text columns and, therefore, unlike regular IEEE
% journals, the available text space is not reduced by their presence.
% If you want to put a publisher's ID mark on the page you can do it like
% this:
%\IEEEpubid{0000--0000/00\$00.00~\copyright~2015 IEEE}
% or like this to get the Computer Society new two part style.
%\IEEEpubid{\makebox[\columnwidth]{\hfill 0000--0000/00/\$00.00~\copyright~2015 IEEE}%
%\hspace{\columnsep}\makebox[\columnwidth]{Published by the IEEE Computer Society\hfill}}
% Remember, if you use this you must call \IEEEpubidadjcol in the second
% column for its text to clear the IEEEpubid mark (Computer Society jorunal
% papers don't need this extra clearance.)

% use for special paper notices
%\IEEEspecialpapernotice{(Invited Paper)}

% for Computer Society papers, we must declare the abstract and index terms
% PRIOR to the title within the \IEEEtitleabstractindextext IEEEtran
% command as these need to go into the title area created by \maketitle.
% As a general rule, do not put math, special symbols or citations
% in the abstract or keywords.
\IEEEtitleabstractindextext{%
\begin{abstract}
Dimensionality Reduction (DR) scatterplot layouts have become a ubiquitous visualization tool for analyzing multidimensional datasets. Despite their popularity, such scatterplots suffer from occlusion, especially when informative glyphs are used to represent data instances, potentially obfuscating critical information for the analysis under execution. Different strategies have been devised to address this issue, either producing overlap-free layouts that lack the powerful capabilities of contemporary DR techniques in uncovering interesting data patterns or eliminating overlaps as a post-processing strategy. Despite the good results of post-processing techniques, most of the best methods typically expand or distort the scatterplot area, thus reducing glyphs' size (sometimes) to unreadable dimensions, defeating the purpose of removing overlaps. This paper presents \textit{Distance Grid (DGrid)}, a novel post-processing strategy to remove overlaps from DR layouts that faithfully preserves the original layout's characteristics and bounds the minimum glyph sizes. We show that DGrid surpasses the state-of-the-art in overlap removal (through an extensive comparative evaluation considering multiple different metrics) while also being one of the fastest techniques, especially for large datasets. A user study with 51 participants also shows that DGrid is consistently ranked among the top techniques for preserving the original scatterplots' visual characteristics and the aesthetics of the final results.
\end{abstract}

% Note that keywords are not normally used for peerreview papers.
}

% make the title area
\maketitle

% To allow for easy dual compilation without having to reenter the
% abstract/keywords data, the \IEEEtitleabstractindextext text will
% not be used in maketitle, but will appear (i.e., to be "transported")
% here as \IEEEdisplaynontitleabstractindextext when the compsoc 
% or transmag modes are not selected <OR> if conference mode is selected 
% - because all conference papers position the abstract like regular
% papers do.
\IEEEdisplaynontitleabstractindextext
% \IEEEdisplaynontitleabstractindextext has no effect when using
% compsoc or transmag under a non-conference mode.

% For peer review papers, you can put extra information on the cover
% page as needed:
% \ifCLASSOPTIONpeerreview
% \begin{center} \bfseries EDICS Category: 3-BBND \end{center}
% \fi
%
% For peerreview papers, this IEEEtran command inserts a page break and
% creates the second title. It will be ignored for other modes.
\IEEEpeerreviewmaketitle

Scatterplots are among the most prevalent visualizations in exploratory data analysis~\cite{8017602, 9226404, 7864468}. Despite their popularity, they suffer from the same issue present in most 3D (we love to hate) visual representations: \textit{occlusion}~\cite{7160096}. As the overlap between visual markers (or glyphs) representing the data instances increases, scatterplots become less effective~\cite{6634128, 8017602, 9226404}, affecting our comprehension~\cite{6484064} and the correctness of analytical tasks, since non-visible objects are prone to be ignored~\cite{10.1145/1101616.1101643}. Even simple design choices like glyphs' rendering order~\cite{7864468} can result in misleading layouts, and the problem is amplified when they convey complex information that occupies more space, such as images~\cite{peximage} or other informative glyphs~\cite{6065026}.

Different strategies have been proposed for the general case to tackle this problem, including sampling, abstraction, re-sizing, changing opacity, and a combination of those~\cite{9226404}. Despite their differences, the underlying common idea is to transform a scatterplot so that typical tasks of correlation estimation, class separation, outlier detection, distribution detection, and point value reading~\cite{7864468} are still valid. For the particular case of Dimensionality Reduction (DR) visualization~\cite{8383983}, this list can be relaxed. Since exact positions in the axes are usually unimportant, especially when non-linear techniques are used, correlation estimation and point-value reading cannot be performed. This allows some flexibility in the glyphs' positions which can be taken advantage of to remove overlaps. 

Some solutions to address this problem can be found in the literature, usually split into overlap-free and overlap-removal strategies. In the former group, techniques seek to create DR scatterplots without overlap~\cite{Pinho2009, Pinho2009_Hexboard} but lack the powerful capabilities of contemporary DR techniques in uncovering interesting data patterns. On the latter, post-processing strategies are devised to remove overlaps of any given scatterplot by rearranging the glyphs while maintaining, as much as possible, the characteristics of the original scatterplot. In this group, optimization of cost functions has been suggested~\cite{Nieto2014, Gansner2010}, with the inherent problems of numerical stability of such solutions. Triangulation~\cite{Nachmanson2016}, orthogonal scan-line-based algorithms~\cite{Misue1995, Hayashi1998, Dwyer2006, garderen2017minimun}, and ``gridfying''~\cite{cutura2022hagrid} approaches are also common, usually delivering better results. In general, however, existing solutions still have problems preserving multiple scatterplot characteristics such as plot dimensions and the glyphs' relative positions, favoring some aspects to the detriment of others and often producing results with glyphs of unreadable dimensions, defeating the purposes of overlap removal. 

This paper presents \textit{Distance Grid (DGrid)}, a novel approach for overlap removal in DR scatterplots. It combines a density-based strategy to generate auxiliary points with a novel space-partitioning method to produce overlap-free layouts that faithfully preserve different characteristics of input scatterplots. Compared to seven state-of-the-art techniques, DGrid presents the best trade-off regarding multiple aspects while producing low distortions and bounding the dimensions of the created layouts, consistently resulting in visual representations with readable glyphs. Also, in a user study with $51$ participants, DGrid was selected as one of the best techniques for preserving the general appearance of original scatterplots while rendering aesthetically pleasant layouts.

In summary, the main contributions of this paper are:

\squishlist
\item  A novel, fast, and highly precise method for overlap removal in DR scatterplots that presents a good trade-off between different aspects of layout preservation; 
\item A thorough analysis of the literature to better formalize the problem of scatterplot characteristics' preservation for overlap removal, consolidating a set of concepts and metrics;
\item An extensive comparative analysis of overlap removal techniques involving eight state-of-the-art approaches plus our proposed technique.
\item The results of a comprehensive user study that identified the preferences of $51$ participants regarding aesthetics, accuracy, and ease of interpretation over the same set of eight approaches plus DGrid.
\squishend

The remainder of this paper is divided as follows. Sec.~\ref{sec:related} discusses the differences between scatterplot overlap removal and sampling concepts and examines the related work. Sec~\ref{sec:method} formalizes the overlap removal problem, consolidates the existing metrics, and presents our method. Sec~\ref{sec:results} presents a comprehensive analysis, comparing DGrid with the state-of-the-art, discusses potential use cases, and finishes with a user study. Finally, Sec.~\ref{sec:limitations} discusses the existing limitations, and Sec.~\ref{sec:conclusions} draws our conclusions.
\section{Related Work}
\label{sec:related}

Occlusion in scatterplots is a well-discussed problem~\cite{7160096} that can make such a popular visual representation less effective~\cite{6634128, 8017602, 9226404}. In the literature, different solutions are presented to address the involved issues, usually \textit{transforming visual markers or glyphs}, for instance, (i) by adjusting their transparency~\cite{Micallef:2017}, sizes~\cite{Li:2009, Jing:2010} or using density and contour plots~\cite{10478445, Tory:2007, Mayorga:2013}, (ii) by carefully \textit{sampling} the glyphs to display~\cite{Chen:2020, Bertini:2006}, or (iii) by re-arranging glyphs' positions to \textit{remove overlaps}. Here we refrain from discussing the existing literature on glyphs' transformation and sampling. Although relevant solutions to improve the visual aspect of scatterplots, they do not solve the problem of removing overlaps, the focus of this paper.

In general, when scatterplots are used to represent Dimensionality Reduction (DR) layouts, the two most common strategies to address occlusion while maintaining all glyphs are: (1) creating overlap-free layouts or (2) eliminating overlaps as a post-processing strategy. Some techniques have been proposed in the former group, such as  IncBoard~\cite{Pinho2009} and HexBoard~\cite{Pinho2009_Hexboard}. Although interesting solutions, they lack the powerful capabilities of contemporary DR techniques to uncover meaningful data patterns. In the latter (i.e., post-processing strategies), some methods remove overlaps by mapping scatterplot points into orthogonal grid cells preserving distances, such as IsoMatch~\cite{Fried2015}, Kernelized Sorting~\cite{Quadrianto2010KernelizedS}, and NMAP~\cite{Duarte2014}. However, grid approaches negatively affect important analytical tasks such as class separation, outlier, and distribution detection~\cite{7864468}, since the empty space is removed and gaps are usually beneficial~\cite{smallmultgaps2017meulemans}. It is also possible to create distance-preserving grids without considering DR layouts as inputs~\cite{Anderson1998, Strong2011, vanKreveld2004, Raisz1934, McNeill2017, Wood2010, Eppstein2013}, but they are out of our scope.

Still, regarding post-processing strategies, a group of techniques heuristically rearrange layouts by moving glyphs while maintaining, as much as possible, the characteristics of a given scatterplot~\cite{chen2020node, MarcilioJr2019}. This paper discusses the problem when the visual area assigned to render a glyph is rectangular (a bounding box). We refrain from discussing other potential shapes, for instance, diamonds~\cite{Meuleman2019}, since comparison among techniques would be impractical once the shape is considered when calculating most layouts' quality metrics (later discussed in Sec.~\ref{sec:problem}).

RWordle~\cite{Strobelt2012} is one example of an overlap-removal technique that uses bounding boxes. It positions glyphs by searching for empty positions using a spiral pattern, with two variants that differ in how the points are processed: RWordle-L orders glyphs along a scan-line, while RWordle-C orders them using distances. RWordle is not precise in preserving essential characteristics of the original layout, such as the similarity relationships among glyphs~\cite{MarcilioJr2019}. It also results in unreasonable running time for dense scatterplots, causing excessive displacement of glyphs and loss of distance relations.

ProjSnippet~\cite{Nieto2014} focuses on distance preservation by maximizing an energy function that accounts for similarity preservation and overlap removal. It has limitations when the original scatterplot has groups of points with high density. In these cases, the technique uses too much space to reduce overlap, often distorting and creating overlap-free layouts with unreadable (tiny) glyphs, defeating the purpose of removing overlaps. One further problem with ProjSnippet is the optimization procedure that sometimes suffers from numerical instability. Another technique, called MIOLA~\cite{GomezNieto2013}, uses mixed-integer quadratic programming to rearrange rectangular boxes in the visual space. MIOLA usually presents more compact layouts after overlap removal. However, it does not preserve well orthogonal ordering and neighborhoods~\cite{GomezNieto2013}.

Another technique that relies on optimization is PRISM~\cite{Gansner2010}. PRISM creates a proximity graph on top of the original layout using triangulation, then minimizes a stress function considering an overlap factor between each pair of nodes in the graph. Scatterplot density also affects its running time, and it suffers from similar problems with optimization instability, sometimes resulting in layouts that are not overlap-free. Triangulation is also used in the GTree method~\cite{Nachmanson2016}, with the triangulation edges used to define a cost function that builds a minimum cost-spanning tree. GTree interactively grows the edges' length of such a tree until there are no more overlaps. It maintains the original scatterplot aspect ratio but moves glyphs excessively, requiring too much space for the final layout and considerably reducing glyphs size~\cite{chen2020node}.

A well-explored strategy to reduce overlap is to process the $x$ and $y$ axes individually. The Push Force-Scan (PFS)~\cite{Misue1995} is one of the pioneers. PFS orders rectangles in the horizontal/vertical direction and applies the minimum movement to remove overlap among subsequent glyphs, focusing on orthogonal ordering preservation. PFS tends to displace glyphs considerably, not preserving the initial layout's aspect ratio~\cite {chen2020node} and adds unnecessary spaces between glyphs. To reduce such spaces, PFS'~\cite{Hayashi1998} moves glyphs horizontally/vertically based on the maximum movement distance from previous glyphs, unlike the sum of previous ones as PFS. The overlap-free layout is improved but presents unnecessary spread and may not maintain the input layout's aspect ratio. Another technique that treats each axis separately is VPSC~\cite{Dwyer2006}. It defines non-overlap constraints for the $x$ and $y$ axes related to how much a glyph needs to be moved in the corresponding direction. The resulting layouts diverge from the original layouts for scatterplots presenting dense groups of points.

Using a different approach that avoids expensive mechanisms of collision detection and handling, Hagrid~\cite{cutura2022hagrid} uses space-filling curves to remove overlaps. The technique first traces a Hilbert curve of level \textit{l} on top of the scatterplot area and discretizes the scatterplot points' coordinates into grid indices considering this curve. Then, the discrete coordinates are transformed into 1D coordinates on the curve, and if a collision occurs when assigning a point to the curve, the point is moved in the curve to find an empty spot. After the points are assigned to 1D coordinates without collisions, these coordinates are transformed into 2D grid indices and back to 2D coordinates. Hagrid is very fast, enabling interactive applications on top of large scatterplots. However, it distorts the original scatterplot aspect ratio and, as with the previous techniques, presents a large displacement and low preservation of global distance relationships, especially for scatterplots presenting dense groups of points.

Finally, ReArrange~\cite{garderen2017minimun} considers pairs of overlapping nodes, one by one, resolving the occlusion problem with the smallest displacement possible while preserving orthogonal order. Like VPSC, ReArrange finds overlaps among glyphs using a line-sweep algorithm and applies overlap removal on $x$ and $y$ axes separately. ReArrange is one of the best-performing approaches, but it distorts the original layout's aspect ratio and increases the layout dimensions, reducing glyphs sizes.

In our technique, aspect ratio preservation and spread are explicitly controlled, producing overlap-free layouts with low distortions and the same dimensions as the original layouts or with a controlled upper-bound, resulting in visualizations with readable glyphs (assuming they are legible on the original scatterplot). Also, distance and neighborhood relationships are precisely maintained, an essential aspect in many DR analytical tasks, without negatively affecting running time.
\section{Method}
\label{sec:method}

Before discussing our solution, we first formalize the problem of overlap removal in scatterplots and consolidate the existing literature about quality metrics.

\subsection{Problem Formulation and Principles}
\label{sec:problem}

In formal terms, given a scatterplot $\mathcal{P}$ composed of $N$ points or glyphs $\mathcal{P}=\{p_1, \ldots, p_N\} \in\mathbb{R}^2$, where ($w_i, h_i$) are the dimensions of the bounding box around $p_i$ and $(x_i, y_i)$ are the top-left corner coordinates of the bounding box,\textit{overlap removal} techniques aim to create a new scatterplot $\mathcal{P'}=\{p'_i, \ldots, p'_N\}\in\mathbb{R}^2$ where the glyphs' superposition are removed and the overall structure of $\mathcal{P}$ is preserved as much as possible. The quality of overlap-free layouts can be defined and measured through different, often conflicting principles and metrics. Following, we present a consolidated list of the most important principles and associated metrics found in the literature, which later are used to evaluate and compare our solution with the state-of-the-art.

\vspace{0.1in}\noindent\textbf{P1 -- Remove glyphs' overlaps.} The resulting scatterplot $\mathcal{P'}$ should present minimum overlap among glyphs, eliminating occlusions that may hide important information. To measure occlusion, we average the pairwise \textit{overlap degree}~\cite{RAMOSGUAJARDO20201} among all graphical glyphs, computing

\begin{equation}\label{eq:overlap}
overlap = \sqrt{\frac{1}{N(N-1)} \sum_i^N\sum_{j \neq i}^N 
\frac{\mathcal{A}(p_i \cap p_j)}
{\min\{\mathcal{A}(p_i), \mathcal{A}(p_j)\}}},
\end{equation}
where $\mathcal{A}(p_i)$ denotes the bounding box area around $p_i$, and the operator $\cap$ returns the intersection of two glyphs. The overlap degree ranges in $[0,1]$ with $0$ denoting an overlap-free layout. 

\vspace{0.1in}\noindent\textbf{P2 -- Preserve glyphs' global and local distances.} Given the typical application of DR layouts for interpreting distance and neighborhood relationships, an overlap-free layout $\mathcal{P'}$ should preserve as much as possible the global distances between glyphs and local neighborhoods presented in the original scatterplot $\mathcal{P}$. To measure global preservation, we use the well-known \textit{stress}~\cite{kruskal1964}, given by 

\begin{equation}\label{eq:stress}
    stress=\sqrt{\frac{\sum_{i<j}^N (||p_i-p_j|| - ||p'_i-p'_j||)^2}
    {\sum_{i < j}^N ||p_i-p_j||^2}}.
\end{equation}
Stress ranges in $[0,\infty)$ with $0$ indicating perfect preservation of distances.
%Notice that to reduce distortions, the transformed layout $\mathcal{P'}$ should be scaled to the same dimensions of the original layout $\mathcal{P}$. 
To measure the local neighborhood preservation, we use the usual \textit{trustworthiness}~\cite{trust2001venna}, given by

\begin{equation}\label{eq:trustwothiness}
trustworthiness = 1 - \frac{2}{NK(2N-3K-1)}\sum_i^N \sum_{j\in U_k^i} (r(i,j) - K),
\end{equation}
where $U_K^i$ is the set of points that are in the neighborhood of size $K$ of $p'_i \in \mathcal{P'}$ but not in the neighborhood of size $K$ of $p_i \in \mathcal{P}$, and $r(i,j)$ is the rank of point $p_j$ in the ordering according to the distance from $p_i$ in the original scatterplot. Trustworthiness ranges in $[0,1]$, the larger, the better, and measures how different is the neighborhood rank of $p'_i \in \mathcal{P'}$ compared to the original neighborhood rank of $p_i \in \mathcal{P}$.

\vspace{0.1in}\noindent\textbf{P3 -- Preserve glyphs' relative positions.} An overlap-free layout $\mathcal{P'}$ should preserve not only distances but also relative positions among glyphs. That is, if $p_i$ is at left/right above/below $p_j$, the same ordering should be observed in $p'_i$ and $p'_j$. This is usually referred to as user mental map preservation and is measured using \textit{orthogonal ordering}~\cite{MISUE1995183}, given by

\begin{equation}\label{eq:orthogonal}
\begin{array}{l}
ordering=\dfrac{1}{N(N-1)} \Bigg( 
\sum_{i,j}^N 
\begin{cases} 
1, & \mbox{if } x_i > x_j \wedge x'_i < x'_j\\ 
0, & \mbox{otherwise} 
\end{cases}
+ \\ 
\qquad\qquad\qquad\qquad\qquad\quad\sum_{i,j}^{N}
\begin{cases} 
1, & \mbox{if } y_i > y_j \wedge y'_i < y'_j\\ 
0, & \mbox{otherwise} 
\end{cases} \Bigg).
\end{array} 
\end{equation} 
The orthogonal ordering ranges in $[0,1]$, with $0$ indicating perfect order preservation. 

\vspace{0.1in}\noindent\textbf{P4 -- Preserve aspect ratio.} The overlap-free representation should preserve the original scatterplot's aspect ratio, reducing distortions. The usual metric to estimate \textit{aspect ratio}~\cite{JGAA-532} is

\begin{equation}
    aspect = \max\left(\frac{W'_{bb} \times H_{bb}}{H'_{bb}\times W_{bb}},  \frac{H'_{bb} \times W_{bb}}{W'_{bb}\times H_{bb}}\right),
\end{equation}
where $W'_{bb}$ and $H'_{bb}$ are the width and height of $\mathcal{P'}$ bounding box, calculated using

\begin{equation}\label{eq:bb}
\begin{split}
W_{bb} = |\max_{i \leq N}(x_i + w_i) - \min_{j \leq N} x_j| \\
H_{bb} = |\max_{i \leq N}(y_i + h_i) - \min_{j \leq N} y_j|
\end{split}
\end{equation}

The \textit{aspect ratio} ranges in $[1,\infty)$, with $1$ the target value. 

\vspace{0.1in}\noindent\textbf{P5 -- Minimize glyphs' displacement.} The translations applied to the glyphs to remove the overlaps and create $\mathcal{P'}$ should be as small as possible to preserve the general appearance of $\mathcal{P}$~\cite{MISUE1995183}. To measure this, we use the average displacement~\cite{4658149}, given by

\begin{equation}\label{eq:displacement}
    displacement=\frac{1}{N \sqrt{W'_{bb} * H'_{bb}}} \sum_i^N ||p_i - p'_i||.
\end{equation}

The \textit{displacement} ranges in $[0,\infty)$, with $0$ indicating no changes in glyphs' positions. Notice that the center of both $\mathcal{P}$ and $\mathcal{P'}$ should be translated to the origin to better capture the relative movement~\cite{JGAA-532}.

\vspace{0.1in}\noindent\textbf{P6 -- Limit glyphs' minimum size.} Finally, the area occupied by the overlap-free representation should be limited to controlling glyphs' minimum size, avoiding creating unreadable layouts and resulting in large empty areas~\cite{JGAA-532}. Visual representations with glyphs that cannot be read defeat the general purpose of increasing layouts' readability by removing overlaps. To quantify this, \textit{layout spread}~\cite{MISUE1995183} can be measured using

\begin{equation}\label{eq:spread}
    spread = \frac{W'_{bb} \times H'_{bb}}{W_{bb} \times H_{bb}},
\end{equation}
where $W_{bb}$ and $H_{bb}$ are the original scatterplot bonding box dimensions calculated using Eq.~(\ref{eq:bb}).

In summary, in addition to the leading goal of removing overlaps (\textbf{P1}), these principles capture two conflicting perspectives, \textit{structure preservation} (\textbf{P2}, \textbf{P3}, and \textbf{P4}) and \textit{readability} (\textbf{P5} and \textbf{P6}). For instance, consider the trivial solution of uniformly scaling the original scatterplot $\mathcal{P}$ by the minimum possible factor that results in an overlap-free layout~\cite{Meuleman2019}. That would result in optimum values for \textbf{P1} (remove overlaps), \textbf{P2} (distances/neighborhoods preservation), \textbf{P3} (orthogonal ordering), and \textbf{P4} (aspect-ratio preservation). However, it would typically present poor results for \textbf{P5} (displacement) and \textbf{P6} (minimum glyphs size) since, for relatively dense scatterplots, the scale factor can be quite large, resulting in substantial amounts of empty spaces and tiny unreadable glyphs.

\begin{figure*}[htb]
    \centering
    \includegraphics[width=\linewidth]{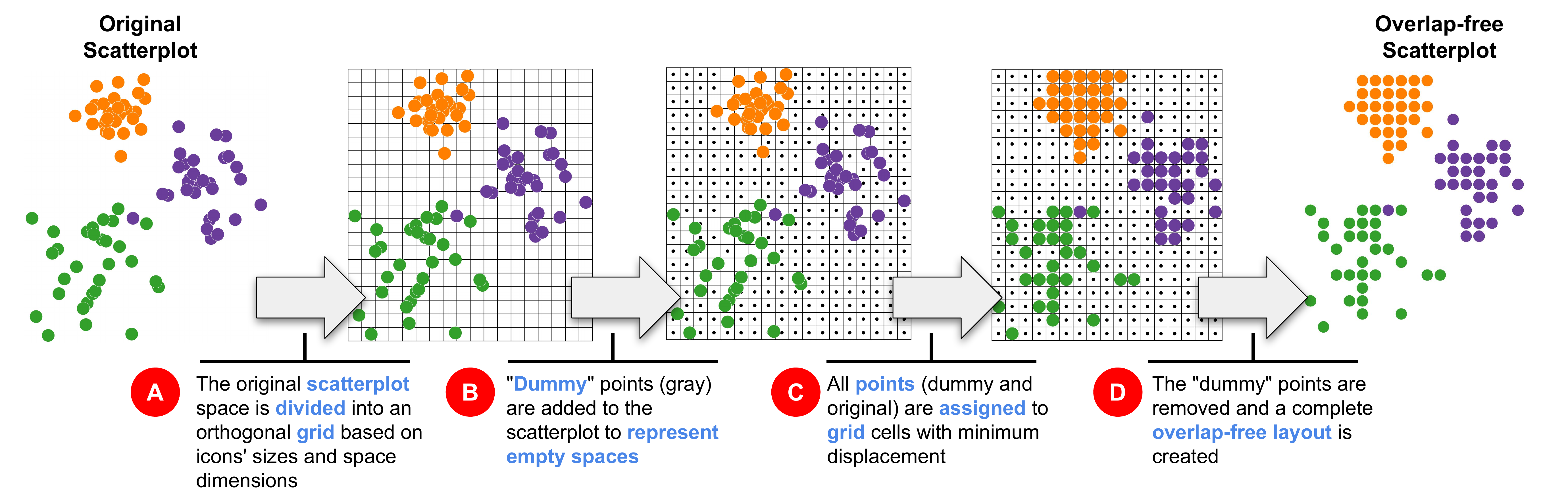}
    \caption{
Overview of DGrid process. The scatterplot area is first split into a grid (A), and ``dummy'' points (small black dots) are crafted to represent empty space (B). Finally, original and ``dummy'' points are assigned to grid cells (C), and the ``dummy'' points are removed (D), resulting in a completely overlap-free layout.}
    \label{fig:overview}
\end{figure*}

\subsection{Distance Grid (DGrid)}
\label{sec:distgrid}

Aiming at finding a good compromise between these perspectives, we present \textit{Distance Grid (DGrid)}, a new solution based on grid assignment to completely remove overlaps in scatterplot visual representations, producing readable layouts while preserving as much as possible, the original scatterplot structures.

DGrid has two major phases divided into different steps (see Fig.~\ref{fig:overview}). In the first, an orthogonal grid $\mathcal{G}$ is superimposed to the scatterplot area with cell sizes equal to the dimensions of the largest bounding box glyph in $\mathcal{P}$, and with the number of rows $R$ and columns $C$ defined by the original scatterplot dimensions (width and height)~\circledr{A}. Then, in the second phase, each scatterplot point is assigned to the closest individual cell so that overlaps are removed. The assignment phase is composed of three different steps. Since usually, the number of cells ($R \times C$) is much larger than the number of scatterplot points $N$ (we later discuss what happens when this does not hold), we first carefully add extra points, called ``dummy'' points, in low-density regions of the scatterplot to represent empty spaces~\circledr{B}. Once the ``dummy'' points are added, the original and ``dummy'' points are assigned to individual grid cells using a grid assignment process that minimizes displacement~\circledr{C}, then the ``dummy'' points are removed~\circledr{D}. The general idea is to use the ``dummy'' points to represent empty regions in the scatterplot so that frontiers between groups and outliers are preserved. The result is a complete overlap-free layout (\textbf{P1}) since each point is assigned to an individual cell, displacing the points as little as possible (\textbf{P5}), thus preserving original distances (\textbf{P2}) and relative positions (\textbf{P3}) while also maintaining the scatterplot aspect ratio (\textbf{P4}) and the original glyph sizes (\textbf{P6}). We start by explaining how to generate the orthogonal grid and the ``dummy'' points.

\subsubsection{Generating the Grid and Adding Dummy Points}

Generating the orthogonal grid $\mathcal{G}$ is straightforward. To limit glyphs' minimum sizes (\textbf{P6}), the cell dimensions are set to $(w_{max}, h_{max})$ where $w_{max}$ and $h_{max}$ are the maximum width and height among all glyphs bounding boxes. Also, to preserve the aspect ratio (\textbf{P4}), $\mathcal{G}$ width and height are defined to be the same as the original scatterplot $\mathcal{P}$, setting the number of columns of $\mathcal{G}$ to $C = \lceil{W_{bb} / w_{max}}\rceil$ and the number of rows to $R = \lceil{H_{bb} / h_{max}}\rceil$. Although fulfilling \textbf{P4} and \textbf{P6}, this only works if the number of cells $R \times C$ is larger than or equal to the number of points $N$. If this is not the case, the result is occluded glyphs, violating \textbf{P1}. To address such situations, the grid's dimensions can be increased by a percentage factor $\Delta \geq 1$, setting

\begin{equation}\label{eq:delta}
\begin{array}{l}
C = \lceil \sqrt{\Delta} \times W_{bb} / w_{max}\rceil\text{, and}\\ 
R = \lceil \sqrt{\Delta} \times H_{bb} / h_{max}\rceil\text{.}
\end{array}
\end{equation}

This preserves the aspect ratio (\textbf{P4}) of the overlap-free layout and potentially improves other structure-preserving metrics (\textbf{P2} and \textbf{P3}). However, it reduces the glyphs' size considering that the final visual representation has the same dimensions of the original scatterplot $\mathcal{P}$, possibly violating \textbf{P6} depending on how large $\Delta$ needs to be defined.

After creating the orthogonal grid, points are assigned to the ``closest'' grid cells in $\mathcal{G}$ to remove the overlaps. Given the nature of the proposed assignment strategy (later discussed in Sec.~\ref{sec:assignment}), if $\mathcal{P}$ is assigned to $\mathcal{G}$ as is, all the points would be grouped in the top-right corner of the final layout. Therefore, to represent empty spaces aiming at preserving the general appearance of the original scatterplot, that is, the existing groups, frontiers between groups, and outliers, ``dummy'' points are added to the list of points to assign to $\mathcal{G}$. The idea is to add the ``dummy'' points in low-density regions but close to original points $p_i$. This way, glyphs' displacement is reduced (\textbf{P5}) since the potential movement of original points is reduced. Also, the represented empty spaces focus on the characteristics of the existing groups of points as much as possible since they are close to or within those groups (borders).

In this process, we first define a list of candidate ``dummy'' points $\mathcal{D}=\{d_1, d_2, \ldots \} \in \mathbb{R}^2$, generating a point per each cell $g_i \in \mathcal{G}$ not occupied by any original point. Since original points will occupy $N$ cells, we select $(R \times C) - N$ points placed in low-density regions from this list. In other words, we remove $N$ ``dummy'' points from the candidate list considering the scatterplot density. To calculate the region density of a point $d_i$, we first count the number of original points lying inside each cell, defining $\mathbf{G}_{R \times C}$, then we convolve a Gaussian kernel of size $M \times M$ with $\mathbf{G}$ setting $d_i$ to its center, where $M$ is calculated as

\begin{equation}\label{eq:mask}
M = \frac{W_{BB} \times H_{BB}}{\sum_i^N (w_i \times h_i)},
\end{equation}
and kernel's $\sigma = (M-1)/6$ (notice that $M$ is rounded to the closest larger odd number). In this way, the kernel size is proportional to the level of detail allowed in the overlap-free layout, which is proportional to the number of empty cells per occupied cell and smoothly covers the region. We also zero-padded $\mathbf{G}$, adding $M/2$ rows and columns to its top/bottom and left/right, respectively.

\begin{algorithm}{}
\renewcommand{\algorithmicrequire}{\textbf{Input:}}
\renewcommand{\algorithmicensure}{\textbf{Output:}}
\algrenewcommand\algorithmicindent{1.0em}%
\begin{algorithmic}
\Require 
  \State $\mathcal{P}=\{p_1,p_2,\ldots p_N\}$ \Comment{Original Scatterplot}
  \State $W_{bb}, H_{bb}$ \Comment{Bounding box of $\mathcal{P}$ (Eq.(\ref{eq:bb}))}
  \State $\Delta \geq 1$ \Comment{Visual space scaling factor. Increase empty space.}
  \State $w_{max}, h_{max}$ \Comment{Max glyph bounding-box width/height in $\mathcal{P}$}
  \State $x_{min}, x_{max}, y_{min}, y_{max}$ \Comment{Max/min glyph coordinates in $\mathcal{P}$}
  \State $C \gets \lceil \sqrt{\Delta} \times W_{bb} / w_{max} \rceil$ \Comment{Calculate the grid number of columns}
  \State $R \gets \lceil \sqrt{\Delta} \times H_{bb} / h_{max} \rceil$ \Comment{Calculate the grid number of rows}
  \vspace{0.25cm}
\Ensure
  \State $\mathcal{D}=\{d_1, \ldots, d_{(R \times C)-N}\}$ \Comment{``Dummy'' points}
  \vspace{0.25cm}
\Function{Dummy}{$\mathcal{P}, C, R$}
    %\State $\mathcal{D} \gets \varnothing$    
    \State $\mathbf{G}_{R \times C} \gets$ \textsc{Count}($\mathcal{P}, C, R$) \Comment{Count the number of points per cell}
    \State $\mathbf{K}_{M \times M} \gets$ \textsc{Mask}($\mathcal{P}$)\Comment{Calculate an $M \times M$ kernel (Eq.(\ref{eq:mask}))}
    \For{$r < R$}
        \State $y \gets r \times (y_{max}-y_{min}) / (R-1) + y_{min}$
        \For{$c < C$}
            \If{$g_{r,c} = 0$} \Comment{Grid cell is empty}
                \State $x \gets c \times (x_{max}-x_{min}) / (C-1) + x_{min}$
                \State $\delta \gets \sum_{i}^{M} \sum_{j}^{M} k_{i,j} \times g_{\lfloor r-(M/2)+j \rfloor, \lfloor c-(M/2)+i \rfloor}$
                \State $\mathcal{D} \gets \mathcal{D} \cup (x,y,\delta)$ \Comment{Add candidate ``dummy'' point}
        \EndIf
        \EndFor
    \EndFor
    \State \Return \textsc{Filter}($\mathcal{D}, (R \times C) - N$)
\EndFunction
\end{algorithmic}
\caption{Creating ``dummy'' points.}
\label{alg:dummy}
\end{algorithm}

If different candidate ``dummy'' points present the same density and we must choose some for the final $(R \times C) - N$ list, we select the closest to any original point. Algorithm~\ref{alg:dummy} details the process of creating the ``dummy'' points. In this algorithm, function {\textsc{Count}($\mathcal{P}, C, R$)} returns a grid $\mathbf{G}_{R \times C}$ with the number of original points per grid cell, {\textsc{Mask}($\mathcal{P}$)} return the Gaussian kernel of size $M \times M$, and {\textsc{Filter}($\mathcal{D}, (R \times C) - N $)} return a list of $(R \times C) - N$ ``dummy'' points placed in low-density regions close to original points.

As already discussed, $R \times C \geq N$, in other words, $M \geq 1$ (Eq.~(\ref{eq:mask})), so all points can be assigned to individual grid cells. Although satisfying this restriction guarantees that an overlap-free layout is produced, the results may degenerate if $M$ is too close to $1$. With $M \gtrapprox 1$, the number of grid cells is too close to the number of original points, so only a few ``dummy'' points are used to represent empty spaces, and the result is a layout without clear borders between groups and outliers. A simple solution is to increase $\Delta$ (see Eq.~(\ref{eq:delta})), which can be interactively done since, in general, DGrid is fast to execute.

\subsubsection{Grid Assignment}
\label{sec:assignment}

The last step in our process is to assign each point $c_i \in \mathcal{C} = \mathcal{P} \cup \mathcal{D}$ (the complete set of original and ``dummy'' points) to an individual grid cell $g_i \in \mathcal{G}$, therefore completely removing the overlaps (\textbf{P1}). For a target grid $\mathcal{G}$ of dimension $(R, C)$ (rows and columns), this process is rather trivial if the marginal distribution of the points $x$-coordinates is perfectly uniform considering a histogram with $C$ bins, and the marginal distribution of the $y$-coordinates is also perfectly uniform but considering a histogram with $R$ bins\footnote{In a perfectly uniform distribution, every bin of the histogram contains the same number of points.}. If such distribution holds, the points are very close to the center of individual grid cells, and the displacement to assign them is minimum (\textbf{P5}). In practice, however, this only holds for overlap-free layouts (or close to that). 

Based on this observation, we devise a new algorithm that recursively bisects $\mathcal{C}$ into non-overlapping partitions until the obtained partitions individually present perfectly marginal uniform distributions. Then, each partition is assigned to a (sub)grid derived from $\mathcal{G}$. The spatial partitioning strategy we use is similar to k-d trees~\cite{kdtree1975bentley}, but instead of only considering point positions to segment the space, it incorporates the grid restriction and translates points to grid cells. Given such constraint, we have $R-1$ different options for horizontal splits and $C-1$ for vertical splits for the first bisection. So the question is how to select the best bisection, or the sequence of horizontal and vertical bisections, that produces the largest partitions with uniform marginal distributions. This is an impractical combinatorial problem to solve. Instead, we split $\mathcal{C}$ approximately in half in the direction (vertical or horizontal) with more rows or columns. In this way, without any expensive test, we increase our probability of getting the largest partitions with marginal uniform distributions -- if the number of points in one partition decreases, this probability increases. However, it reduces such a probability for the other partition. Therefore, half is a natural trade-off. 

\begin{figure}[htb]
\centering
\includegraphics[width=.75\linewidth]{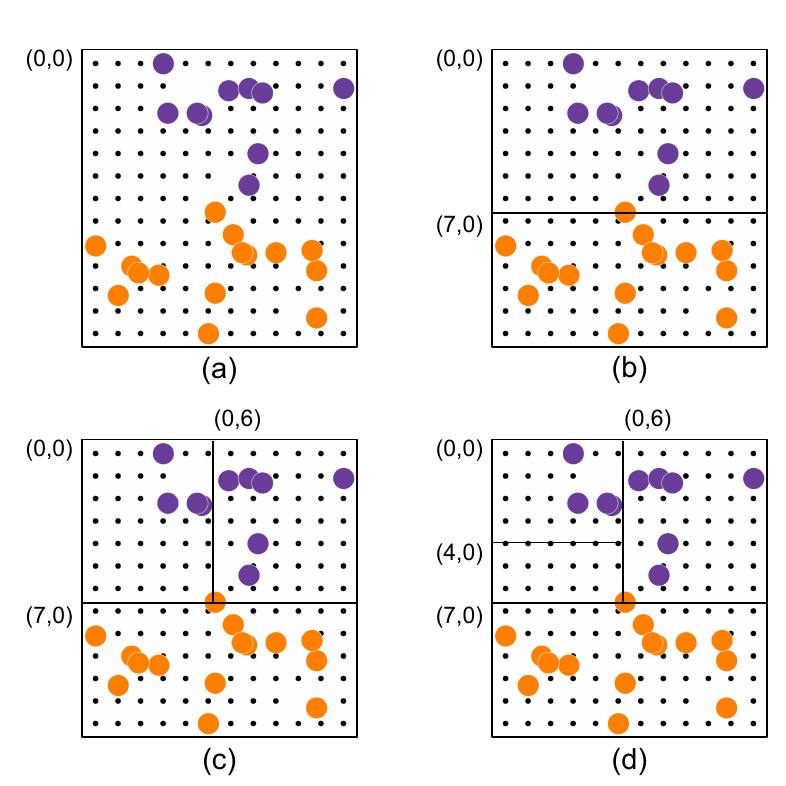}
\caption{Process of assigning points to grid cells. In this example, a scatterplot with $156$ ``dummy'' and original points (a) is assigned to a grid with $13$ rows and $12$ columns. Starting from a horizontal bisection (b), the scatterplot is recursively split until each partition contains only one point, resulting in the grid cell indexes.}
\label{fig:bisect}
\end{figure}

In our process, if $R > C$, we split $\mathcal{C}$ horizontally, obtaining two partitions $\mathcal{C}=\mathcal{C}_1\cup\mathcal{C}_2$, so that the top partition $\mathcal{C}_1$ contains enough points to fill (approximately) half of the target grid, that is, $|\mathcal{C}_1| = \lceil{R/2}\rceil \times C$. Otherwise, we split $\mathcal{C}$ vertically so that the left partition $\mathcal{C}_1$ contains enough points to fill (approximately) half of the target grid, that is, $|\mathcal{C}_1| = R \times \lceil{C/2}\rceil$. This bisecting process is recursively applied to the resulting partitions until obtaining partitions with marginal uniform distribution. This bisecting process results in the minimum displacement for partitions with perfect uniform marginal distributions. Therefore, instead of penalizing our approach's running time by adding the uniform constraint test, we execute the bisecting process until each partition contains only one point, rendering a much faster and simpler process to implement.

The last piece to discuss is how to calculate the grid cell indexes (positions) to which the points in $\mathcal{C}$ are assigned. Without loss of generality, if the first bisection vertically splits $\mathcal{C}=\mathcal{C}_1\cup\mathcal{C}_2$, so that $|\mathcal{C}_1| = R \times \lceil{C/2}\rceil$, by construction, the index of the most-left column to which points in $\mathcal{C}_1$ will be assigned is $0$, and the index of the most-left column to which points in $\mathcal{C}_2$ will be assigned is $\lceil{C/2}\rceil$. Similarly, for a horizontal split, the index of the most-top row to which points in $\mathcal{C}_1$ will be assigned is $0$, and the index of the most-top row to which points in $\mathcal{C}_2$ will be assigned is $\lceil{R/2}\rceil$. In the general case, when an input partition is assigned to a grid with $R'$ rows and $S'$ columns, if $(i,j)$ is the index of the top-left corner cell of the input grid, the index of the top-left corner cell to which points in $\mathcal{C}_1$ will be assigned is $(i,j)$, and the index of the top-left corner cell to which points in $\mathcal{C}_2$ will be assigned is $(i+\lceil{R'/2}\rceil,j)$ for a horizontal cut, and $(i,j+\lceil{C'/2}\rceil)$ for a vertical cut. If the input partition has only one point, $(i,j)$ is the cell index to which the single point is assigned. 

Fig.~\ref{fig:bisect} outlines this process. In this example, a scatterplot with $25$ original points and $131$ ``dummy'' points (small black dots) is assigned to a grid with $R=13$ rows and $C=12$ columns. To start this process, we set the top-left corner cell index to $(0,0)$ (Fig.~\ref{fig:bisect}(a)). Since the input grid has more rows than columns, the first bisection is horizontal (Fig.~\ref{fig:bisect}(b)). The resulting top partition contains (approximately) half of the rows ($7$), and the top-left corner cell of the resulting (sub)grid receives the index $(0,0)$. The bottom partition contains the remaining rows ($6$), and the top-left corner cell receives the index $(7,0)$. Next, the top partition is bisected (Fig.~\ref{fig:bisect}(c)). Since the (sub)grid resulting from this partition has more columns than rows, it is vertically split. Again, each resulting partition receives half of the columns ($6$). The top-left corner cell of the resulting grid from the left partition receives the index $(0,0)$, whereas the top-left corner cell of the resulting grid from the right partition receives the index $(0,6)$. This partitioning process is then recursively applied to the resulting partitions (Fig.~\ref{fig:bisect}(d)) until each partition contains only one point.

Algorithm~\ref{alg:dgrid} puts all these pieces together. Function \textsc{Split}$_y(\mathcal{C}, K)$ performs the horizontal bisection. In this process, $\mathcal{C}$ is sorted according to the $y$-coordinates ($\mathcal{C}$ is viewed as a list). The first $K$ points are assigned to $\mathcal{C}_1$ and the remaining to $\mathcal{C}_2$. The function \textsc{Split}$_x(\mathcal{C}, K)$ performs the vertical bisection using the same process but sorting $\mathcal{C}$ according to the $x$-coordinates. Notice that, from an implementation perspective, we use a pre-sorting strategy~\cite{Brown2015kdtree} so that the $x$- and $y$-coordinates are only sorted once. Although we have discussed our process as a grid index assignment, the depicted algorithm already transforms such indexes into coordinates, multiplying the indexes by the maximum glyphs bounding box $(w_{max}, h_{max})$.

\begin{algorithm}{}
\renewcommand{\algorithmicrequire}{\textbf{Input:}}
\renewcommand{\algorithmicensure}{\textbf{Output:}}
\algrenewcommand\algorithmicindent{1.0em}%
\begin{algorithmic}
\Require 
  \State $\mathcal{C} = \mathcal{P}\cup \mathcal{D}$ \Comment{Scatterplot with ``dummy'' and original points}
  \State $R, C$ \Comment{Grid dimensions (rows and columns)}
  \State $w_{max}, h_{max}$ \Comment{Max glyph width/height in $\mathcal{P}$}
  \vspace{0.25cm}
\Ensure
  \State $\mathcal{P'} = \{p'_1, \ldots p'_N\}$ \Comment{Transformed points}
  \vspace{0.25cm}
\Function{DGrid}{$\mathcal{C}$, $(R,C)$}
    \State \textsc{DGrid\_aux}($\mathcal{C}, (R,C), (0,0)$)
    \State \Return $\mathcal{P'}$
\EndFunction
\vspace{0.25cm}
\Function{DGrid\_aux}{$\mathcal{C}$, $(R,C)$, $(i,j)$}
  \If{$\mathcal{C} \neq \emptyset$} 
    \If{$|\mathcal{C}| = 1$} \Comment{$\mathcal{C}$ has one point}
      \If{$c \in \mathcal{C}$ is an original point}
        \State $\mathcal{P'} \gets \mathcal{P'} \cup (j \times w_{max}, i \times h_{max})$
      \EndIf
    \Else     
      \If{$R > C$}
        \State $\mathcal{C}_1,\mathcal{C}_2 \gets$ \textsc{Split}$_y$($\mathcal{C}$, $\min(|\mathcal{C}|,\lceil{R/2}\rceil\times C)$)                
        \State \textsc{DGrid\_aux}($\mathcal{C}_1$, $(\lceil{R/2}\rceil, C), (i,j)$)
        \State \textsc{DGrid\_aux}($\mathcal{C}_2$, $(R-\lceil{R/2}\rceil, C), (i+\lceil{R/2}\rceil, j)$)
        %\State \Return $\mathcal{C}_1 \cup \mathcal{C}_2$   
      \Else
        \State $\mathcal{C}_1,\mathcal{C}_2 \gets$ \textsc{Split}$_x$($\mathcal{C}$, $\min(|\mathcal{C}|, R \times\lceil{C/2}\rceil$)
        \State \textsc{DGrid\_aux}($\mathcal{C}_1, (R, \lceil{C/2}\rceil), (i,j)$)
        \State \textsc{DGrid\_aux}($\mathcal{C}_2, (R, C-\lceil{C/2}\rceil), (i, j+\lceil{C/2}\rceil)$)
        %\State \Return $\mathcal{C}_1 \cup \mathcal{C}_2$      
      \EndIf
    \EndIf 
  \EndIf
  %\State \Return $\mathcal{C}$
\EndFunction
\end{algorithmic}
\caption{Assigning points to grid cell positions.}\label{alg:dgrid}
\end{algorithm}

Notice that since each original and ``dummy'' point is assigned to an individual cell, and we have the same number of points and cells, the result is a complete overlap-free representation (\textbf{P1}).

\subsubsection{Computational Complexity}

To set DGrid's computational complexity, we first split the suggested process into two major phases, ``dummy'' points creation and grid assignment. Considering $T = R \times C > N$ the total number of grid cells, the complexity of creating the ``dummy'' points is O($T \log T$) if a nearest neighbor data structure is used to find the nearest neighbors, such as a k-d tree~\cite{kdtree1975bentley}, and an O($n \log n$) sorting algorithm is applied to define the final list of points (\textsc{Filter()} function). For the second phase, if an O($n \log n$) algorithm is also used to sort the points' $x$- and $y$-coordinates, its computational complexity is O($T \log T$). Assuming that the original scatterplot occupies a finite area and that the ratio between this area and the summation of all glyphs area is bounded by a small constant (see Eq.~(\ref{eq:mask})), DGrid's overall complexity is O($N \log N$).
\section{Results}
\label{sec:results}

In this section, we present different examples to highlight the advantages of using an overlap-free layout if compared to traditional DR layouts, quantitatively evaluate and compare DGrid against the state-of-the-art, and finish with a user evaluation to measure subjective elements that are not possible to capture through quality metrics.

\subsection{Use-Cases}

To illustrate the benefits of using overlap-free layouts produced by DGrid, we present different examples in this section. The first, presented in Fig.~1, shows a t-SNE~\cite{maaten2008visualizing} projection and layouts produced using DGrid varying the area allocated to the visual space. In this example, the dataset is the \textit{UCI ML Breast Cancer Wisconsin}~\cite{uci2013lichman} composed of $569$ samples of \textit{malignant} and \textit{benign} cancer imaging diagnosis.

\begin{figure*}[ht]
    \centering
\subfigure[Original]{\includegraphics[width=.2\linewidth]{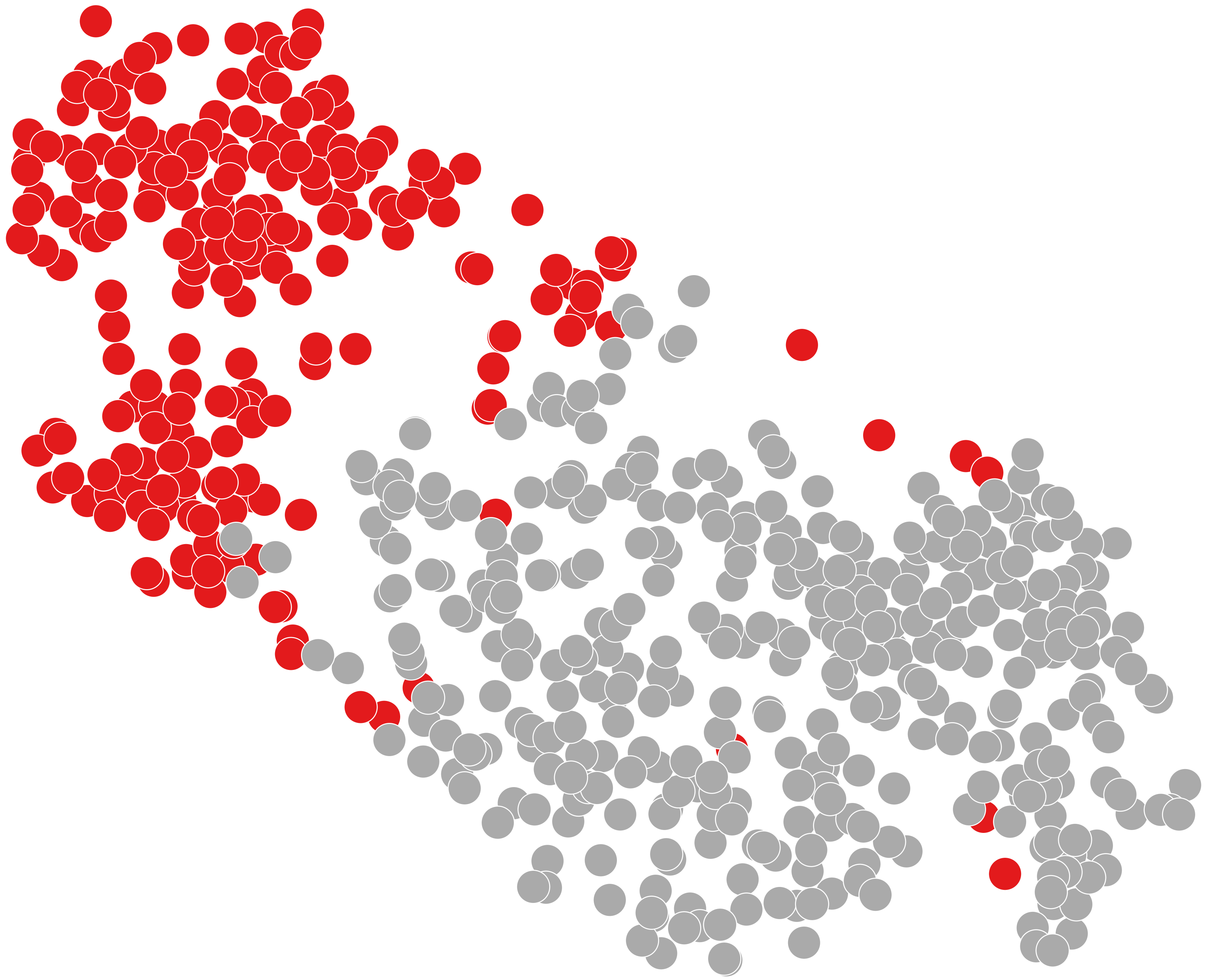}}\qquad
\subfigure[$\Delta=1.0$]{\includegraphics[width=.2\linewidth]{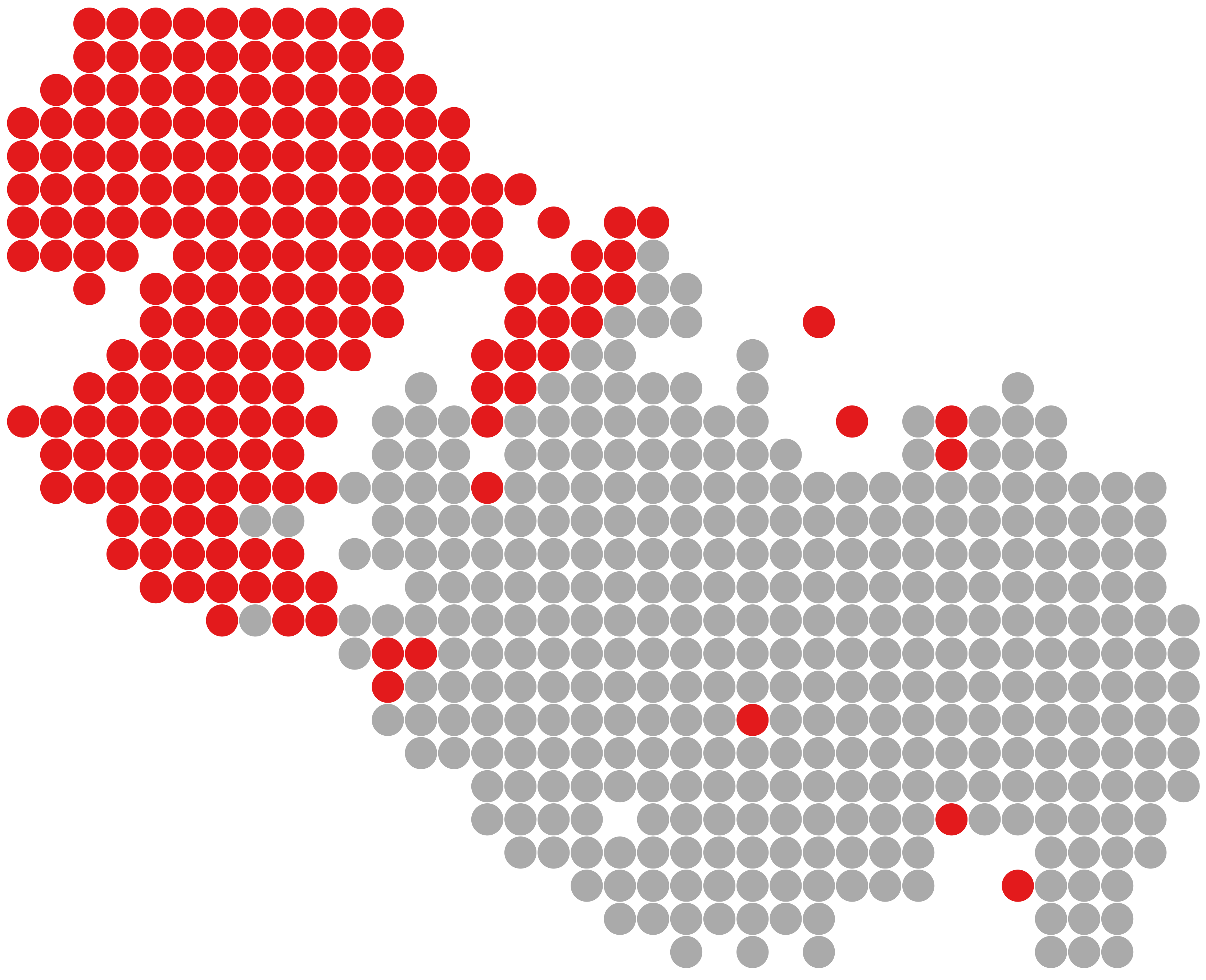}}\qquad
\subfigure[$\Delta=1.25$]{\includegraphics[width=.2\linewidth]{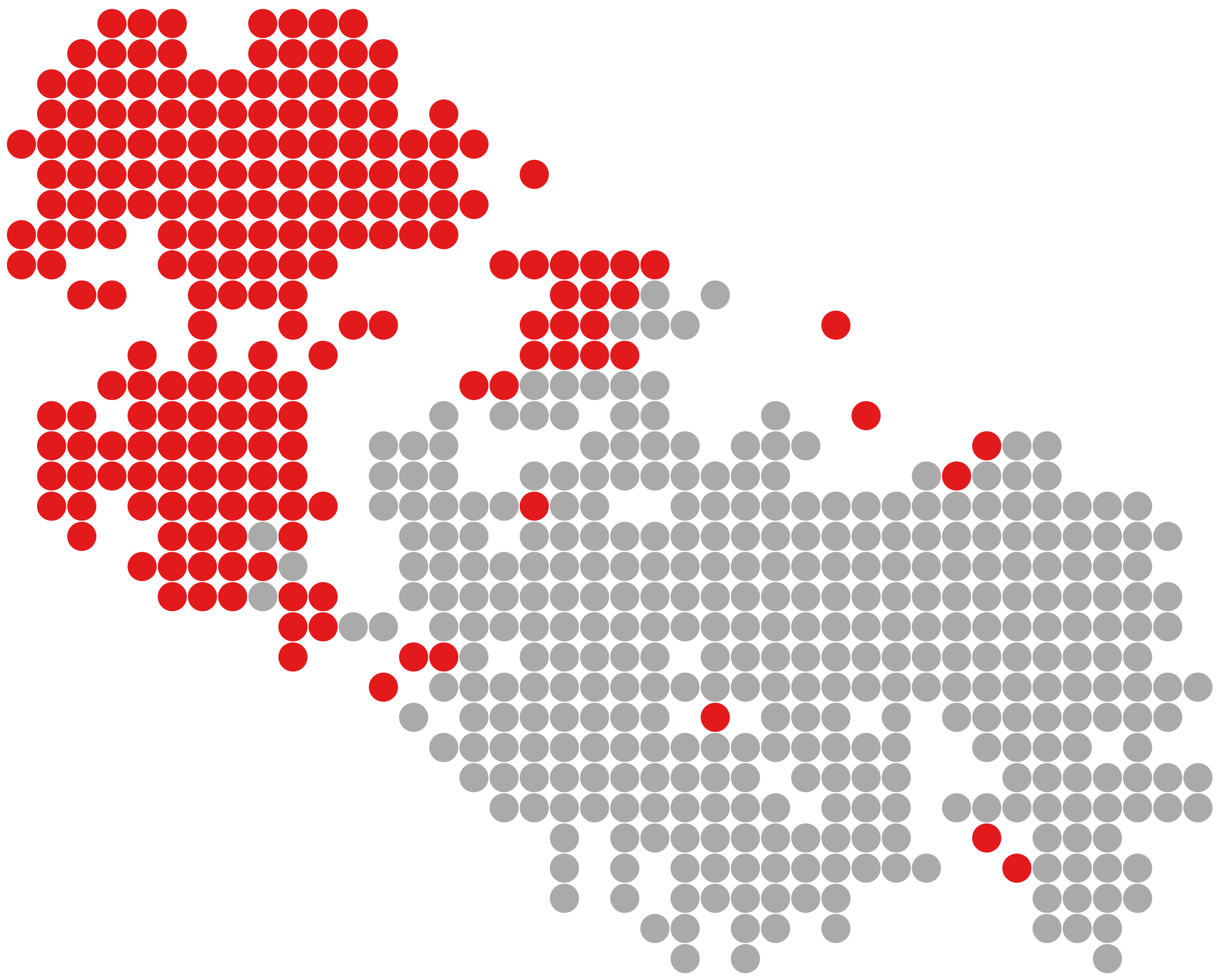}}\\
\subfigure[$\Delta=1.5$]{\includegraphics[width=.2\linewidth]{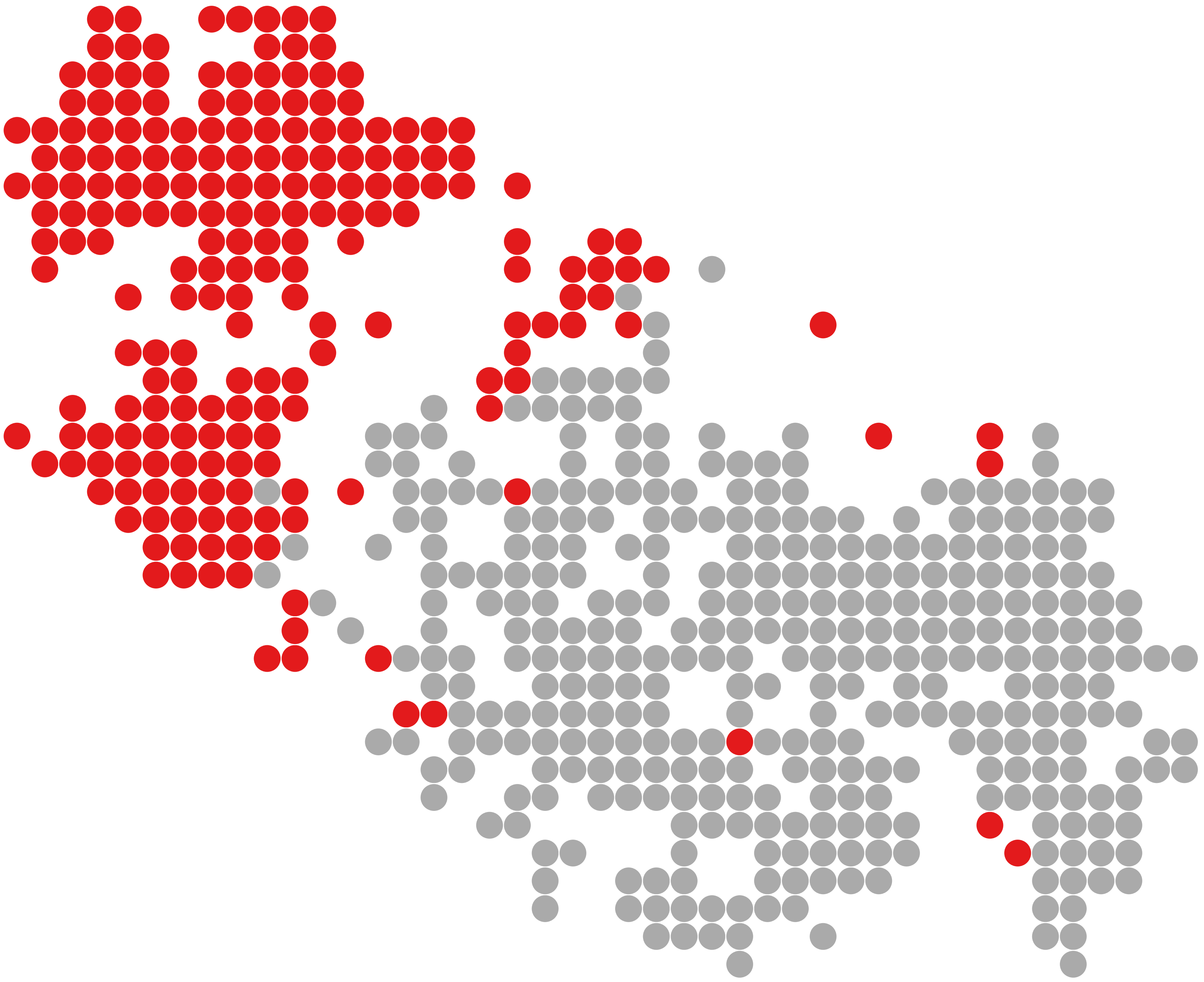}}\qquad
\subfigure[$\Delta=1.75$]{\includegraphics[width=.2\linewidth]{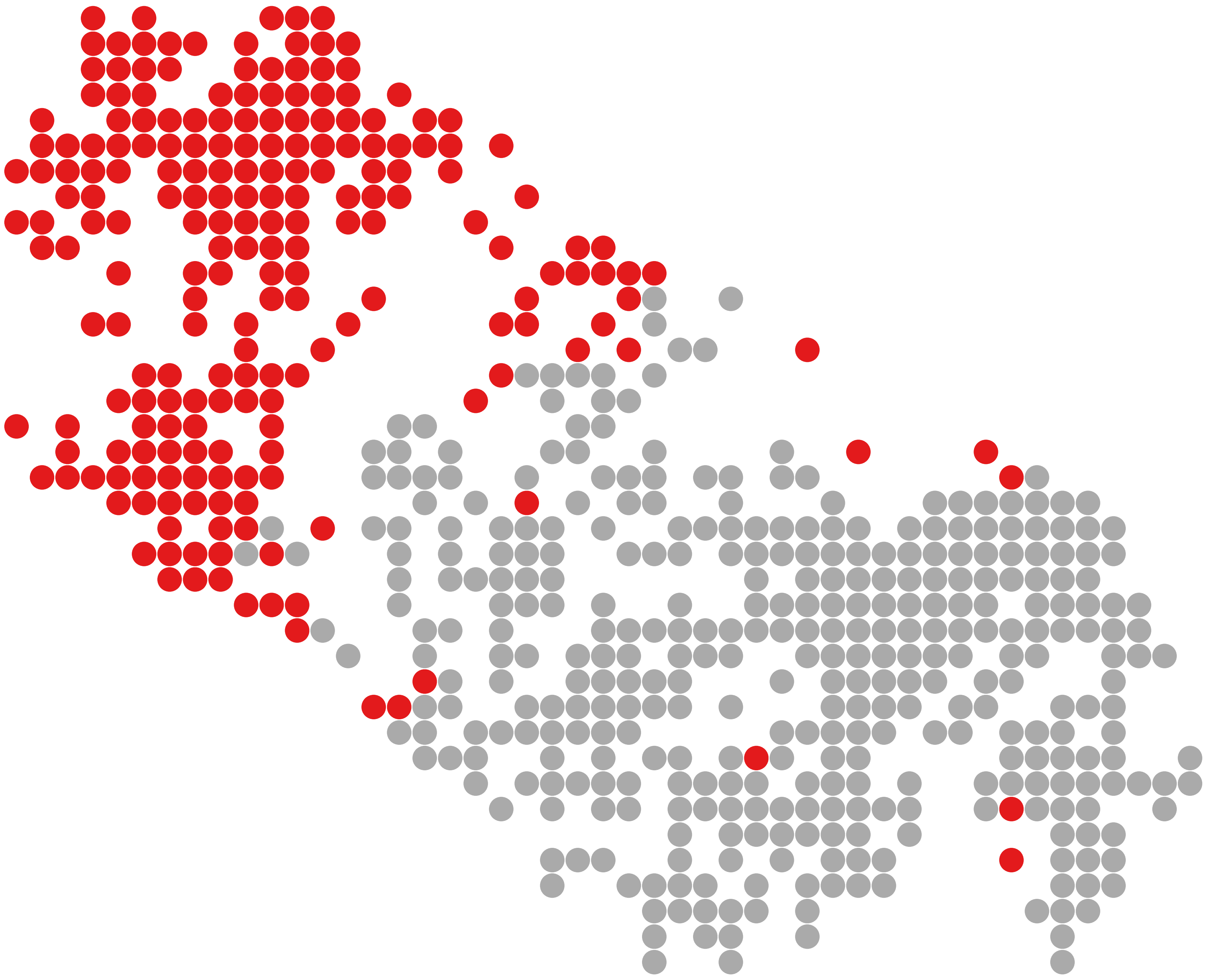}}\qquad
\subfigure[$\Delta=2.0$]{\includegraphics[width=.2\linewidth]{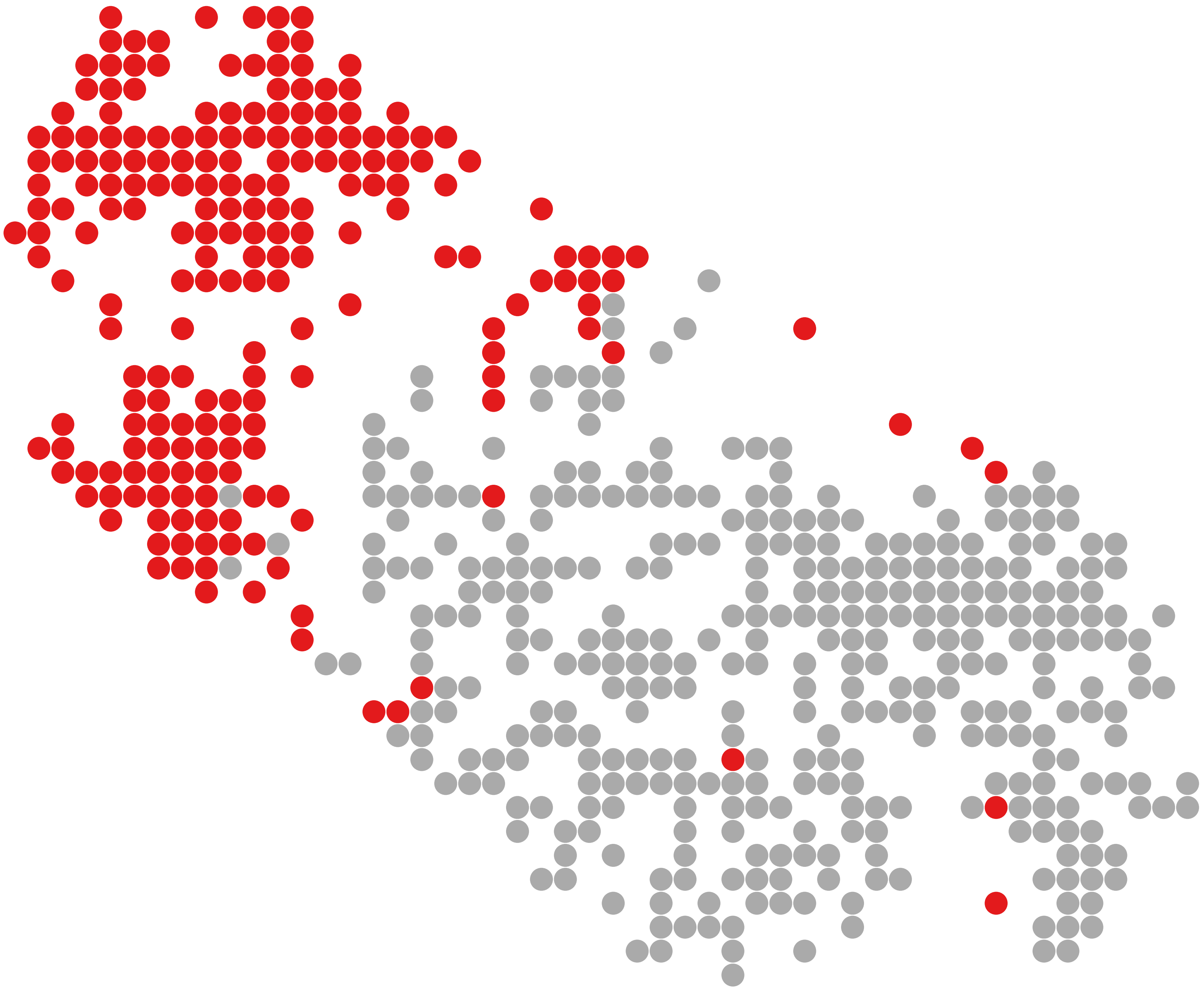}}
\caption{Overlap removal varying the space allocated for the visual representation (or reducing glyphs' size). Given that red points are rendered before gray points, some are occluded due to overplotting (a). After applying DGrid, such points become visible, but the frontiers between groups and sub-groups are lost (b). As the visual area is increased (c-f), multiplying it by a factor ($\Delta$), the original appearance of (sub)group separation is recovered, resulting in a readable overlap-free layout.}
\label{fig:teaser}
\end{figure*}

In the original layout produced by t-SNE (Fig.~1(a)), some of the malignant samples (in red) are occluded by benign samples (in gray) since the dataset is ordered by label, and the red points are rendered before the gray points. When DGrid is applied, these points become visible. Observe the now visible red point, a malignant sample close to the center of gray points (Fig.~1(b)), completely occluded on the original layout. This outlier has practically the same feature values as some benign cancer samples and is very different from most malignant ones. By observing only the original layout, this fact could probably be ignored. Despite the benefits, since the space allocated to the overlap-free layout is the same as the original layout ($\Delta=1.0$), the frontiers between groups and sub-groups are blurred. So, fine differences that could be spotted in different sub-groups are lost. This problem can be mitigated by increasing the visual area, in other words, by reducing glyphs' sizes (Fig.~1(c)-(f)). When the area is $50\%$ larger than the original layout ($\Delta=1.5$), those frontiers become noticeable, and the separation between groups and sub-groups emerges. When the area is doubled ($\Delta=2.0$), most of the structure of the original layout is captured. However, glyphs become noticeably smaller, and user participation in deciding the correct balance between details and glyph sizes becomes necessary. Once DGrid is fast, with interactive rates for datasets with a few thousand instances, on-the-fly experimentation during data analysis is possible.

\begin{figure*}[ht]
    \centering
    \subfigure[Original UMAP layout.]{\includegraphics[width=\textwidth]{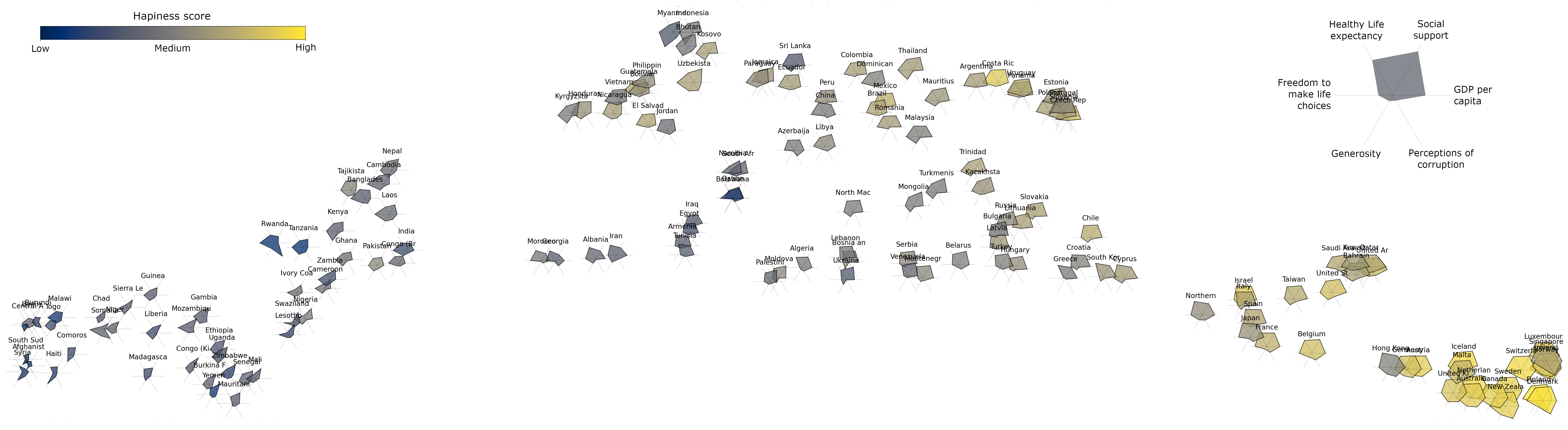}}
    \subfigure[Overlap-free layout.]{ \includegraphics[width=\textwidth]{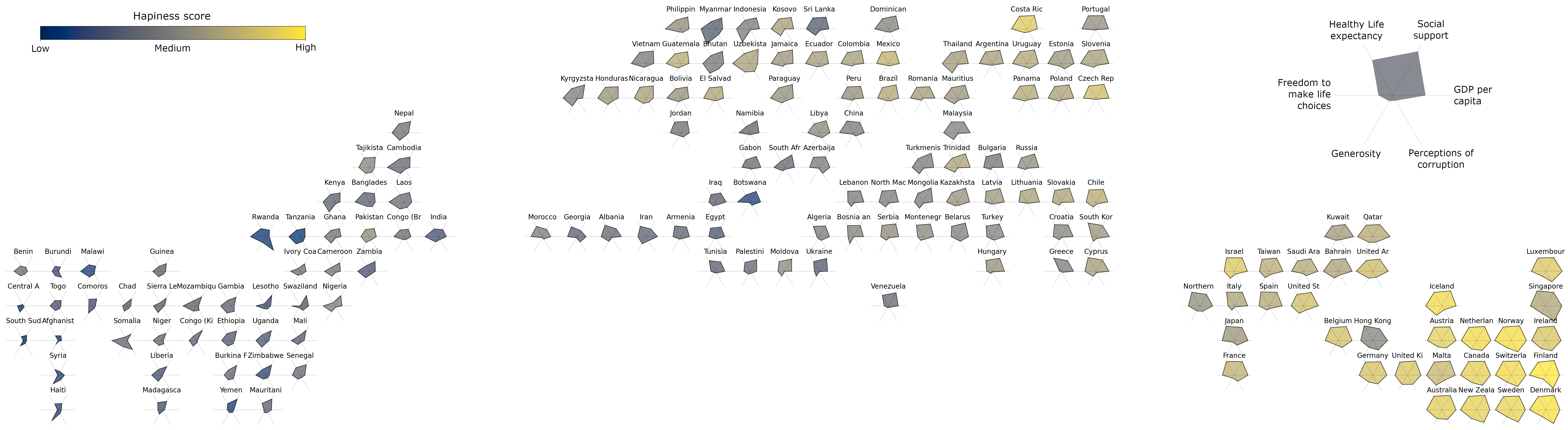}}
    \caption{Original and DGrid overlap-free layouts of a projection from the World Happiness Report 2019 data~\cite{helliwell2019world}. Glyphs encode the perceived happiness as color, and the other six variables are mapped to a starglyph. Even for a small dataset, overlaps affect intra-group analysis and a detailed view of the individual instances in the original layout (a). On the overlap-free representation, intra-group analysis is enhanced, allowing detailed examination of individual instances and existing patterns, helping users take advantage of DR layouts combined with richer glyphs.}
  \label{fig:happiness}
\end{figure*}

Beyond the usual color-coded circular glyphs, the benefits of using richer glyphs have been discussed in the literature as a means to join the quick overview afforded by DR layouts with the ability of glyph-based visualizations to convey information about the original data dimensions~\cite{8967136}. In these cases, the benefit of using overlap-free layouts is even more evident. Fig.~\ref{fig:happiness} presents such an example where \textit{starglyphs}~\cite{6875973} are used to encode the original multidimensional information. This example uses the \textit{World Happiness Report 2019}~\cite{helliwell2019world} dataset~\footnote{\url{https://www.kaggle.com/unsdsn/world-happiness}}. This dataset presents a ranking of $156$ countries based on a score representing how happy their citizens perceive themselves and other six variables to support the explanation of the happiness variation across countries, including GDP per capita, social support, healthy life expectancy, freedom to make life choices, generosity, and perception of corruption.

We project this dataset using the UMAP~\cite{mcinnes2020umap} technique considering these six variables. Fig.~\ref{fig:happiness}(a) presents the original overlapped layout (a high-resolution version of this image is included in the supplemental material). Glyphs encode the happiness score as color, and the country names are written on the top of each glyph. In this layout, three different groups of countries are noticeable, one with high happiness scores, another with low scores, and a transition group composed of countries between these two. Only a few glyphs are ``readable''. Most are difficult to understand without an auxiliary strategy (e.g., zoom). Here we use transparency, a usual solution when simple glyphs are used, to mitigate some overlap problems, but it does not help much. In this example, it is difficult to extract any other information beyond group formation, even for this considerably small dataset. 

The same projection after removing overlaps with DGrid is presented in Fig.~\ref{fig:happiness}. Even after reducing empty space, the three groups are still noticeable, and many other details are visible. In each one of the groups, the general magnitude of the variables follows the same pattern, although it is possible to see that this is not uniform (we use Euclidean distance, so this is expected). And when checking individual countries, some interesting findings can be derived. For example, generosity varies without following the happiness pattern. Finland (bottom-right), for instance, has a low level of perceived generosity but a high level of happiness, while Haiti (bottom-left) presents the opposite behavior. However, we also have Iceland with high generosity and high happiness and Senegal with low generosity and low happiness. A different observation can be made considering GDP per capita, where usually the richer countries present more significant happiness scores. Of course, with some exceptions, like Costa Rica, a not-rich country with high happiness, and Hong Kong, a rich but with a low score. Many other findings can be extracted from this layout, made possible by using more informative glyphs and removing overlaps, leveraging DR layouts to be valuable beyond clustering/segmentation tasks, and allowing the detailed inspection of individual data instances.

\begin{figure*}
    \centering
    \includegraphics[width=\linewidth]{dgrid-cnn-analysis-v2.pdf}
    \caption{Using dimensionality reduction and DGrid to analyze CNN results. The original layout is produced using UMAP, and the instances are colored according to ground truth (A). After applying DGrid to remove overlaps, the general appearance is maintained, but class-outliers are easier to spot (B). One group is zoomed in for analysis, and images are used as glyphs, allowing the visualization of intra-group patterns (C). To further investigate misclassifications, SHAP is used to show which part of the images contributes to the classification mistakes (D).}
    \label{fig:image-analysis}
\end{figure*}

Our last example shows the benefit of using DGrid to process larger datasets, presenting how it can improve Convolutional Neural Networks (CNN) analysis based on DR layouts. In this example, we project the test set with $10,000$ images from the Fashion MNIST~\cite{Xiao2017} dataset using the dense layer as features. The network we use has two convolutional layers, followed by a dense layer with $128$ neurons, followed by a softmax layer with $10$ neurons. Fig.~\ref{fig:image-analysis} shows the original UMAP layout on the left and the DGrid overlap-free on the right. The circles representing the data instances are color-coded according to the ground truth labels of the images, with red points representing the CNN misclassifications.

Some interesting global insights (inter-group) can be derived from these visual representations. For instance, this particular classifier is precise for ``trousers'' (light purple) and ``bags''(light green) since those groups are clearly separated from the others. However, it has difficulties distinguishing between ``coats'' (green), ``shirts'' (gray), and ``pullovers'' (blue), given the high incidence of error and the unclear frontiers between these groups. Also, this classifier considers ``sandals'' (pink), ``sneakers'' (pink), and ``ankle bots'' (yellow) similar; however, it successfully differentiates between them, with most of the errors in the frontiers between groups.

Although such observations can be made using both the original and the over-free layouts, the magnitude of the error is arguably more apparent in the overlap-free version, and rendering order issues are avoided. However, the advantages of an overlap-free layout emerge when class-outlier analysis (intra-group) is executed. The class-outlier analysis is an essential task in classification models and focuses on assessing instances of a given class (colors in this example) inside a homogenous group of another class~\cite{doi:10.1177/1473871617713337}. For this analysis, notice the two highlighted clusters in the original DR layout. Fig.~\ref{fig:image-analysis}~\circledr{A} shows a zoom-in of a subcluster of ``bags'' images (in green). In the overlap-free representation, it is possible to see some class-outlier examples and misclassifications that are hidden in the original layout. In this sub-cluster, two instances from the ``dress'' class (in orange) are far away from their ground-truth cluster, indicating that, although the CNN model was able to classify them correctly, these images are very similar to some ``bags''. This is confirmed by checking the original images, where it is possible to verify the similarity between these two dresses and nearby bags. Also, notice the close misclassified instance (in red), a dress classified as a bag. To help understand why the classifier has confused this sample, we use SHAP~\cite{Lundberg2017} to explain which part of the image contributed more to the final classification. It is possible to notice that the intensity distribution makes it look like a purse and that the predicted class (``Bag'') presents a high contribution (reddish colors) to the full extent of the image, indicating that this CNN may have problems when classifying this particular type of dress.

Finally, Fig.~\ref{fig:image-analysis}~\circledr{B} highlights another example that DGrid makes it easier to understand and look for class-outliers. In this case, we select a ``sandal'' misclassified as an ``ankle boot'' placed in the middle of the well-defined group of ``ankle boots'' (yellow). In other words, it is a sandal image very similar to ankle boots from the classifier perspective. SHAP indicates the high intensity, and the heels contribute to the misclassification. Nearby ankle boot images are shown for comparison, and in Fig.~\ref{fig:image-analysis}~\circledr{C}, two samples from the ``Sandal'' class are presented for illustration. It is interesting to notice that, when looking for (real) similar images on the Internet, it is possible to find the definition of ``Gladiator Strappy Ankle Boots Heels Sandals.'' So a challenging image to even define its ground truth.
\subsection{Evaluation and Comparison}
\label{sec:quant}

For the quantitative evaluation, we compare DGrid against ReArrange, PFS', Hagrid, VPSC, RWordle-L, RWordle-C, ProjSnippet, and PRISM techniques (see Sec.~\ref{sec:related}). In this comparison, we use the $7$ different quality metrics defined in Sec.~\ref{sec:problem} to assess the readability and structure preservation of the resulting overlap-free scatterplots. We did not compare against MIOLA since we could not replicate the code or the results of the original paper with public information.

For the tests, we generate $1,000$ different scatterplots varying the sizes between $500$ and $1,000$ points, the densities, that is, the ratio between scatterplots and glyphs' area (see Eq.(\ref{eq:mask})) in $\{3,5,7,9,11\}$, the aspect ratios between $[1,4]$, and the number of groups in $\{1,2,3,4,5\}$. To generate such groups, points coordinates are generated using different Gaussian distributions allowing to control the density of each group and the level of overlap among groups. We prefer to use synthetic scatterplots since this allows us to control the experiments better, producing layouts with varying aspects and helping to uncover the reasons for the attained results. All the results were generated in an Intel(R) Core(TM) i7-8700 CPU @ 3.20 GHz, 32 GB RAM, Ubuntu 64 bits, NVIDIA GeForce GTX 1660 Ti 22 GB. The techniques were implemented in Java. For the ReArrange, RWordle-C, and RWordle-L, we used the authors' original codes, and we implemented the others -- our Hagrid implementation was based on the author's javascript implementation, and the produced results were identical. We also implemented DGrid in Python, and the code with all the examples presented in this paper is publicly available\footnote{ \url{https://github.com/fpaulovich/dimensionality-reduction}}.

\begin{figure*}[htb]
    \centering
    \includegraphics[width=.195\linewidth]{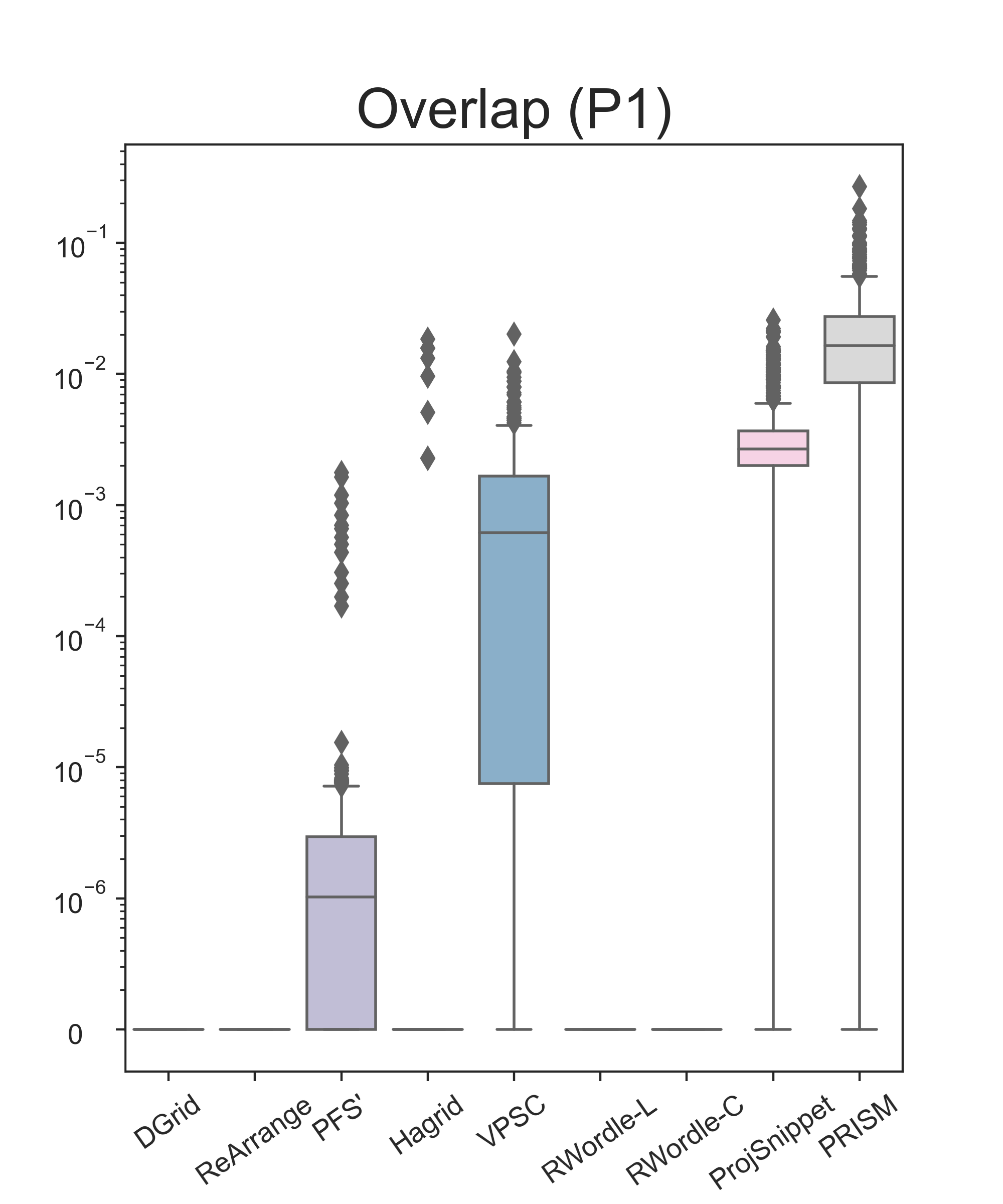}
    \includegraphics[width=.195\linewidth]{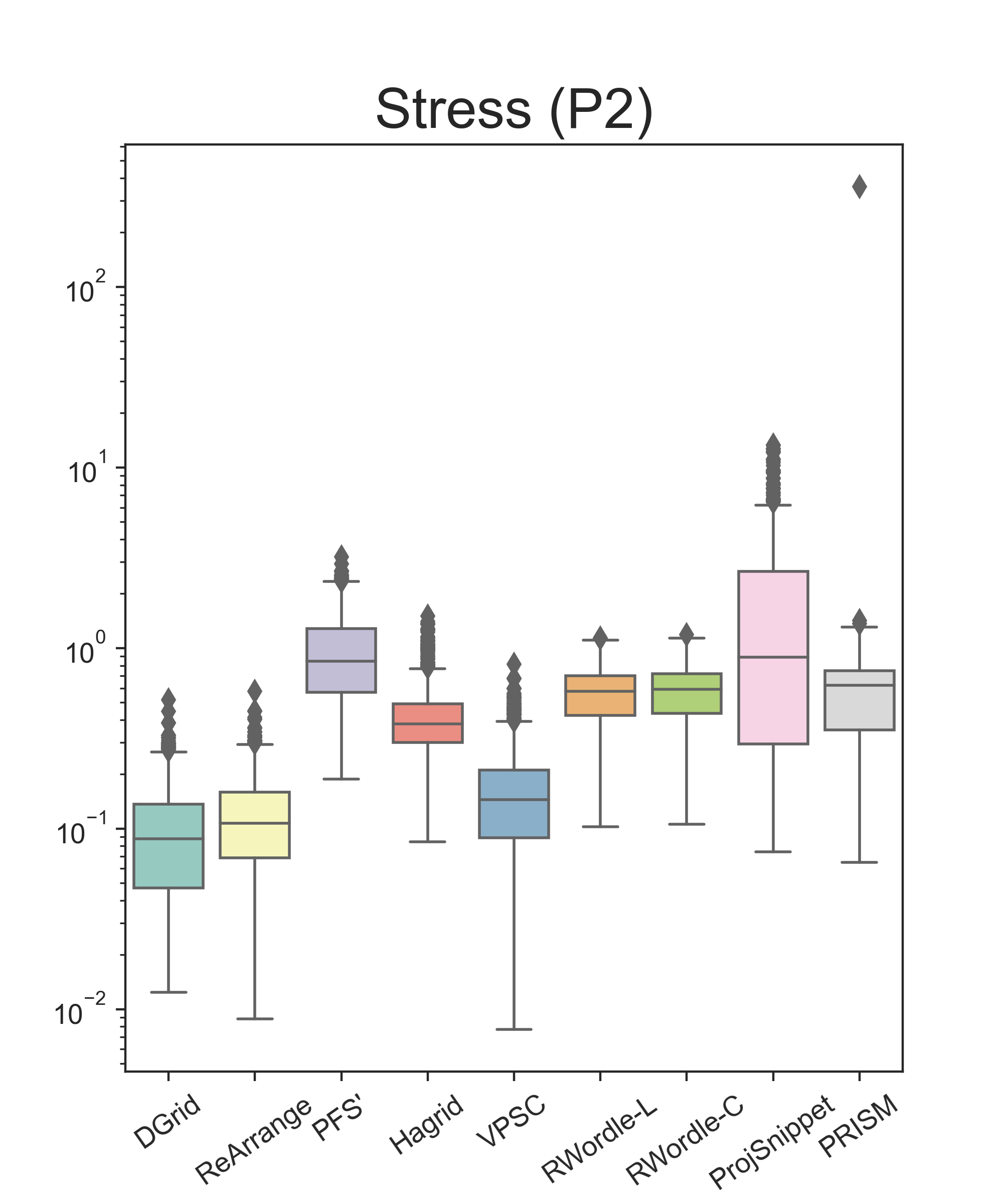}
    \includegraphics[width=.195\linewidth]{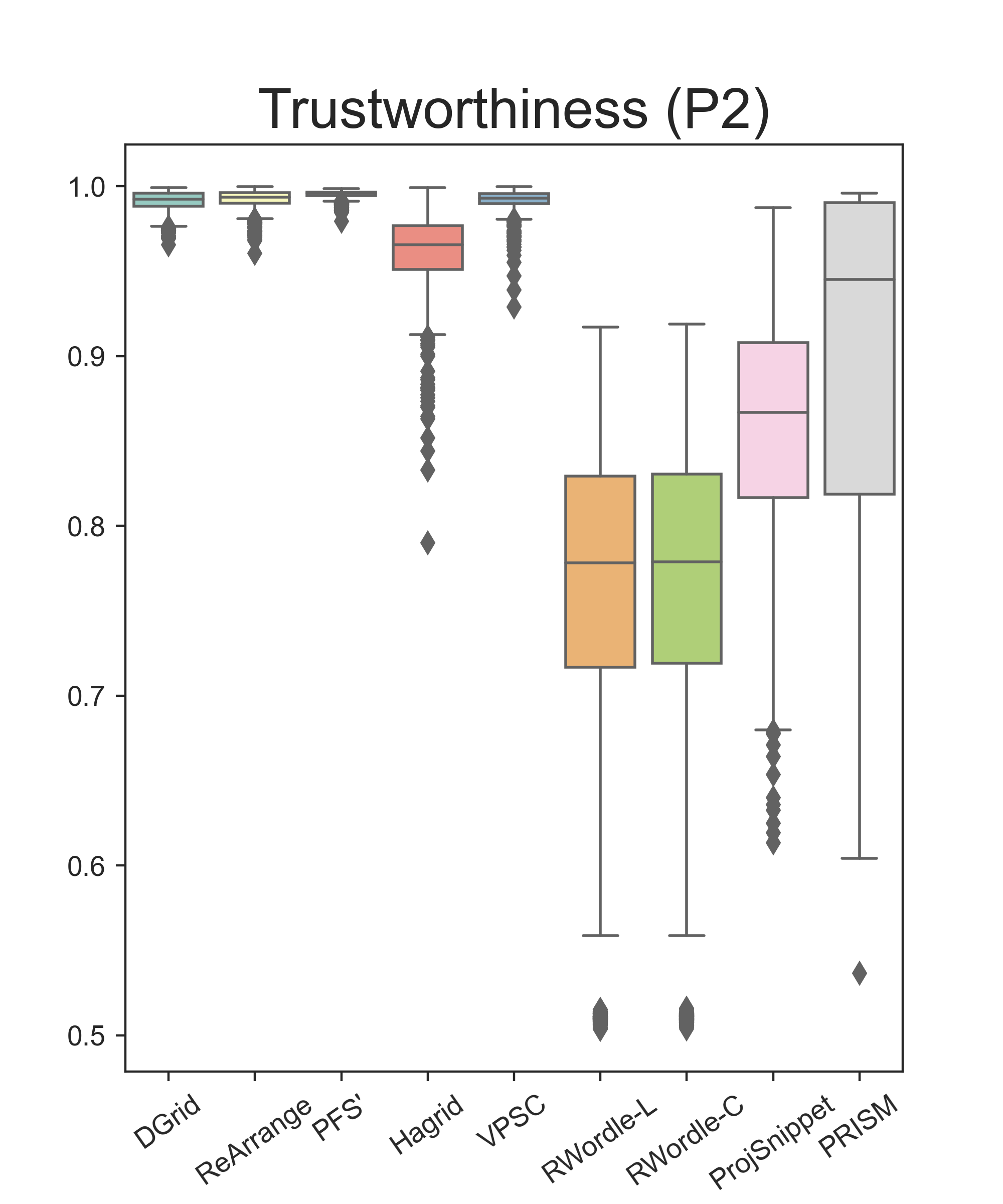}
    \includegraphics[width=.195\linewidth]{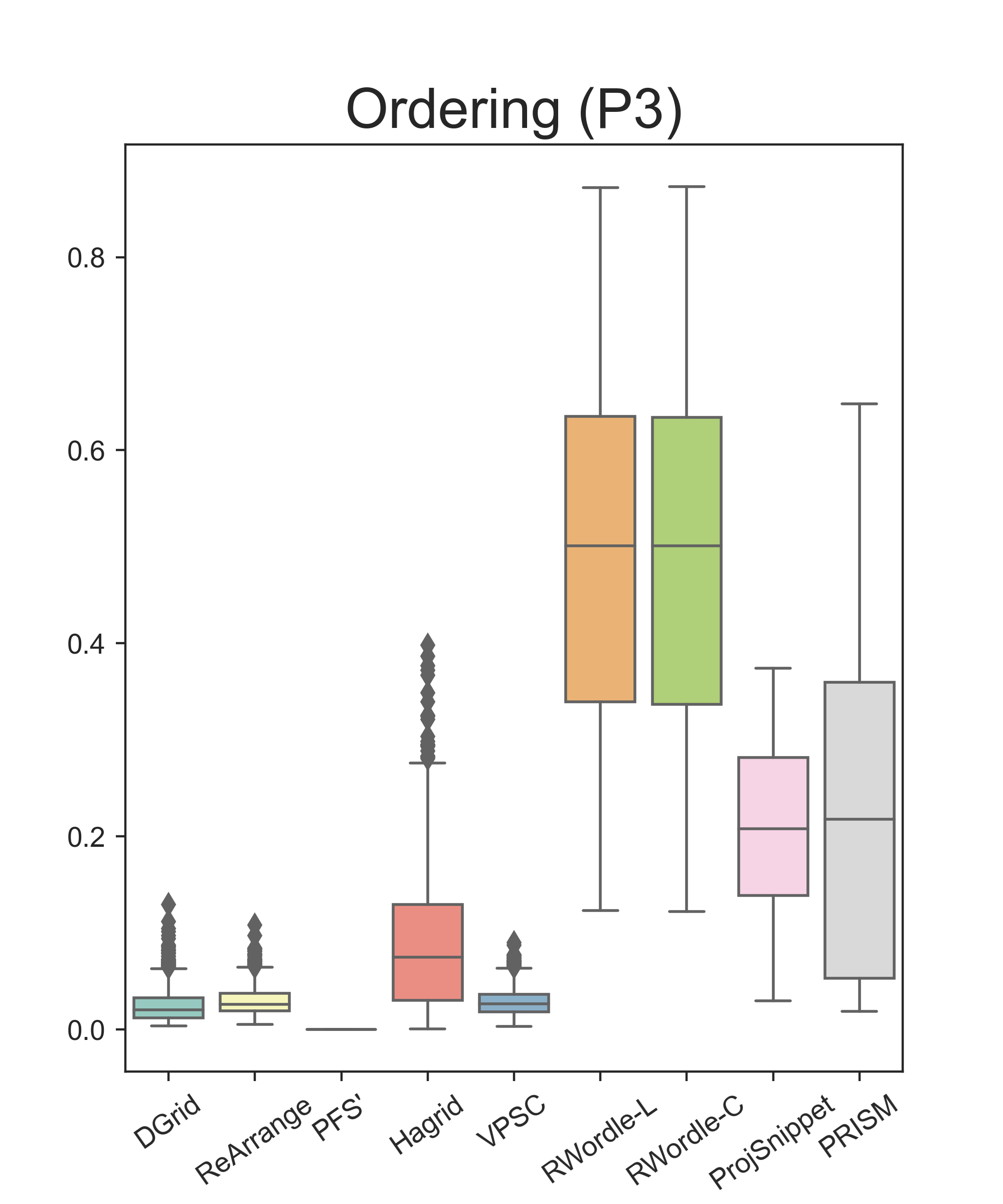}
    \includegraphics[width=.195\linewidth]{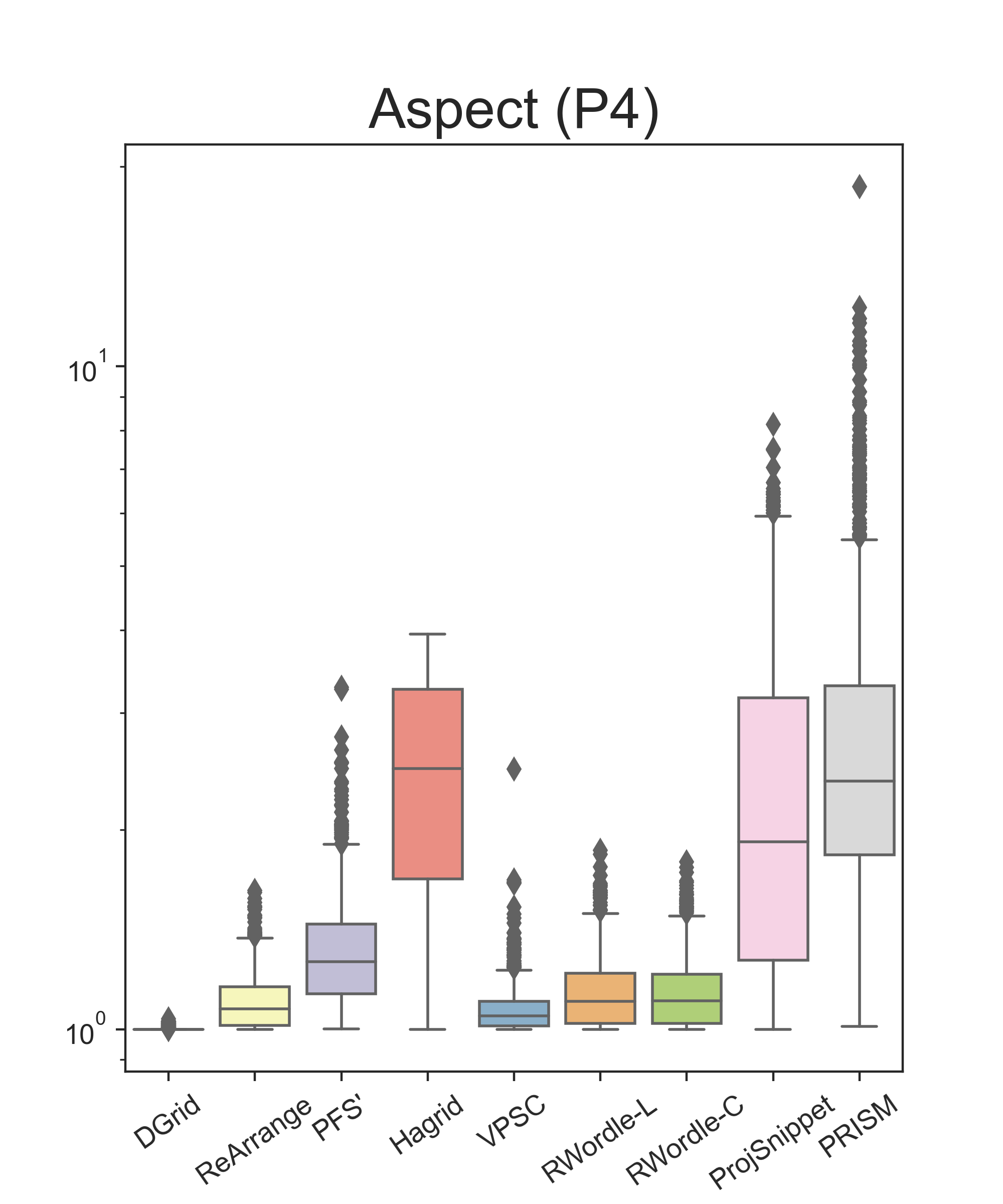}
    \\
    \includegraphics[width=.195\linewidth]{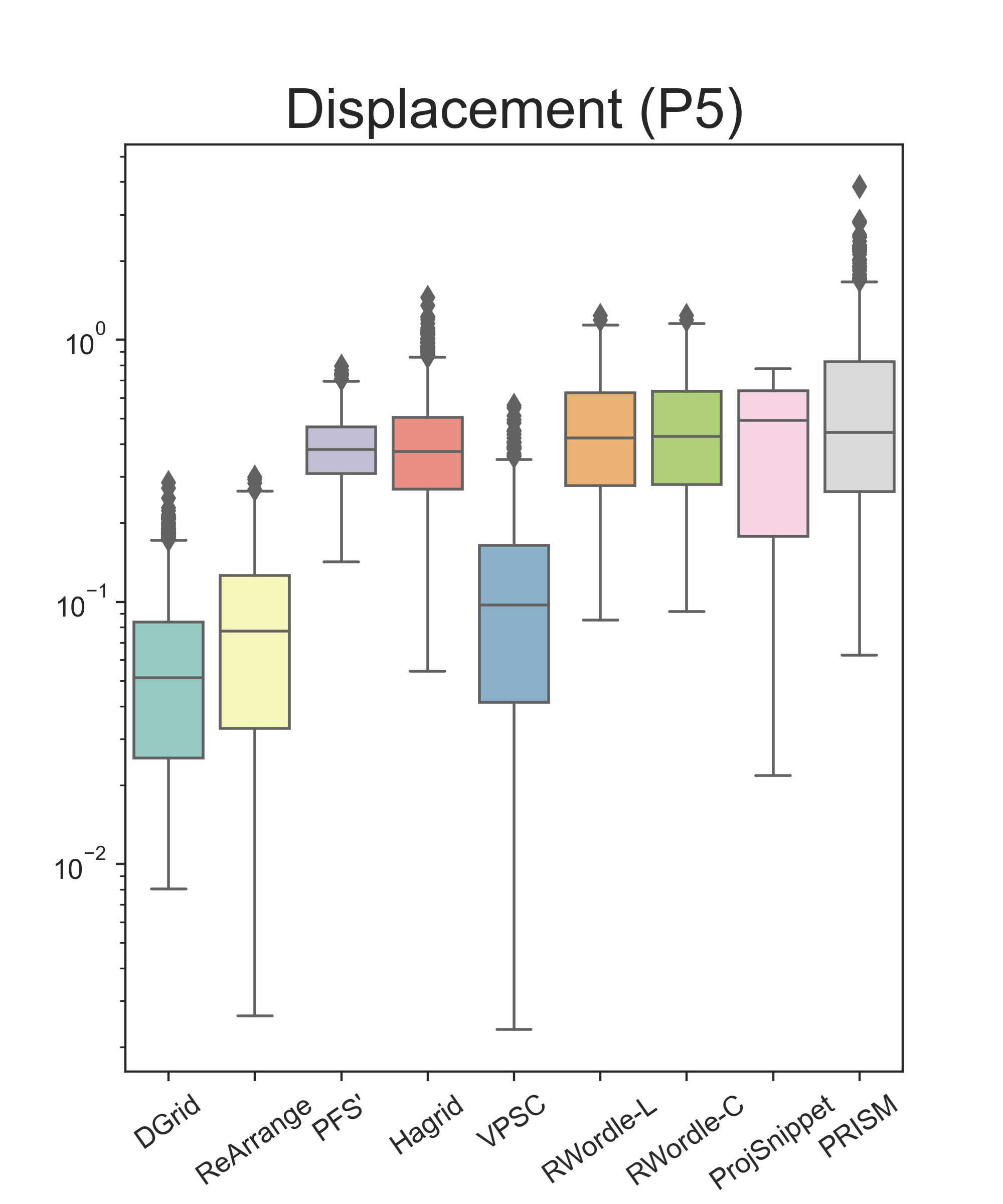}
    \includegraphics[width=.195\linewidth]{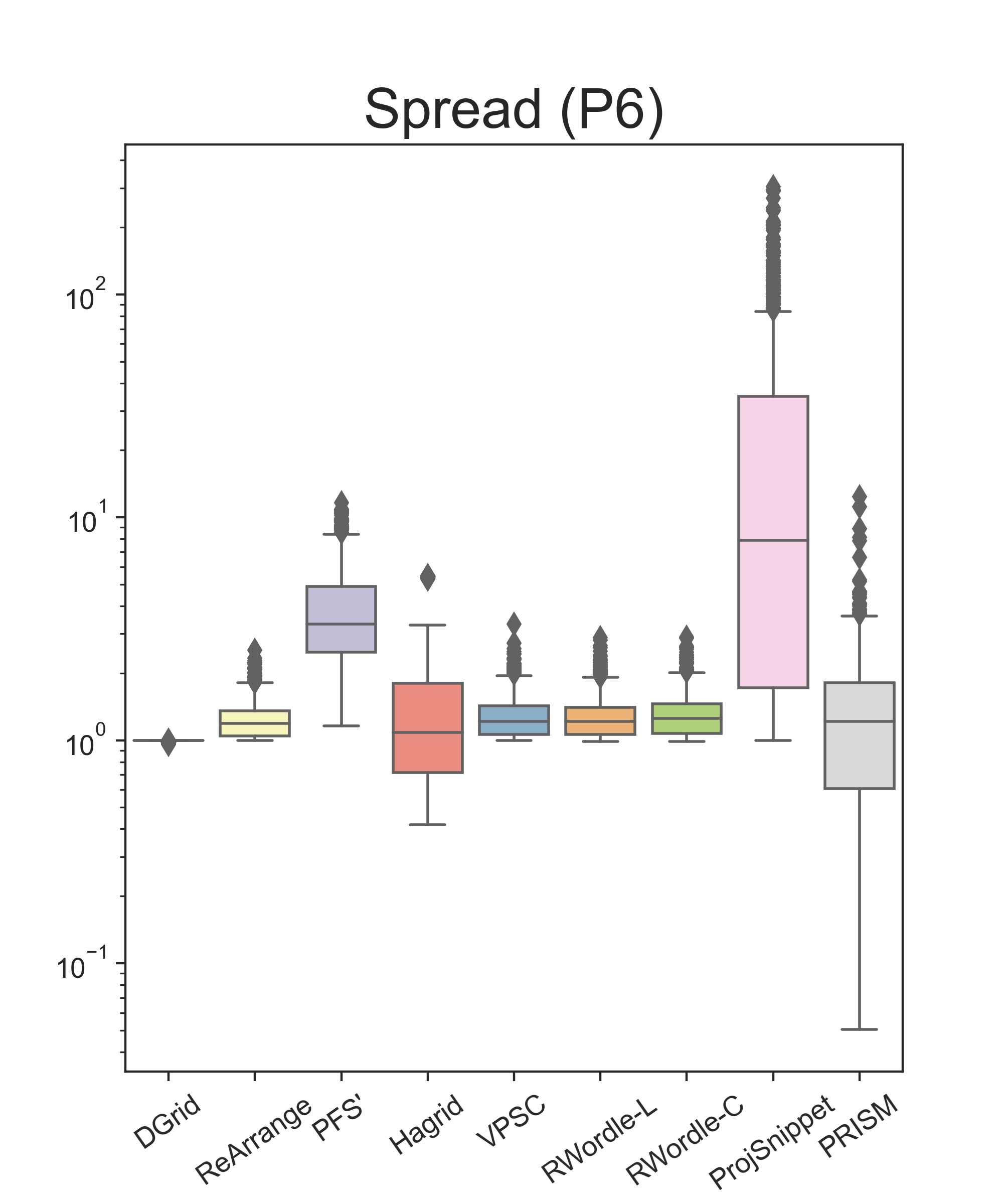}
    \includegraphics[width=.195\linewidth]{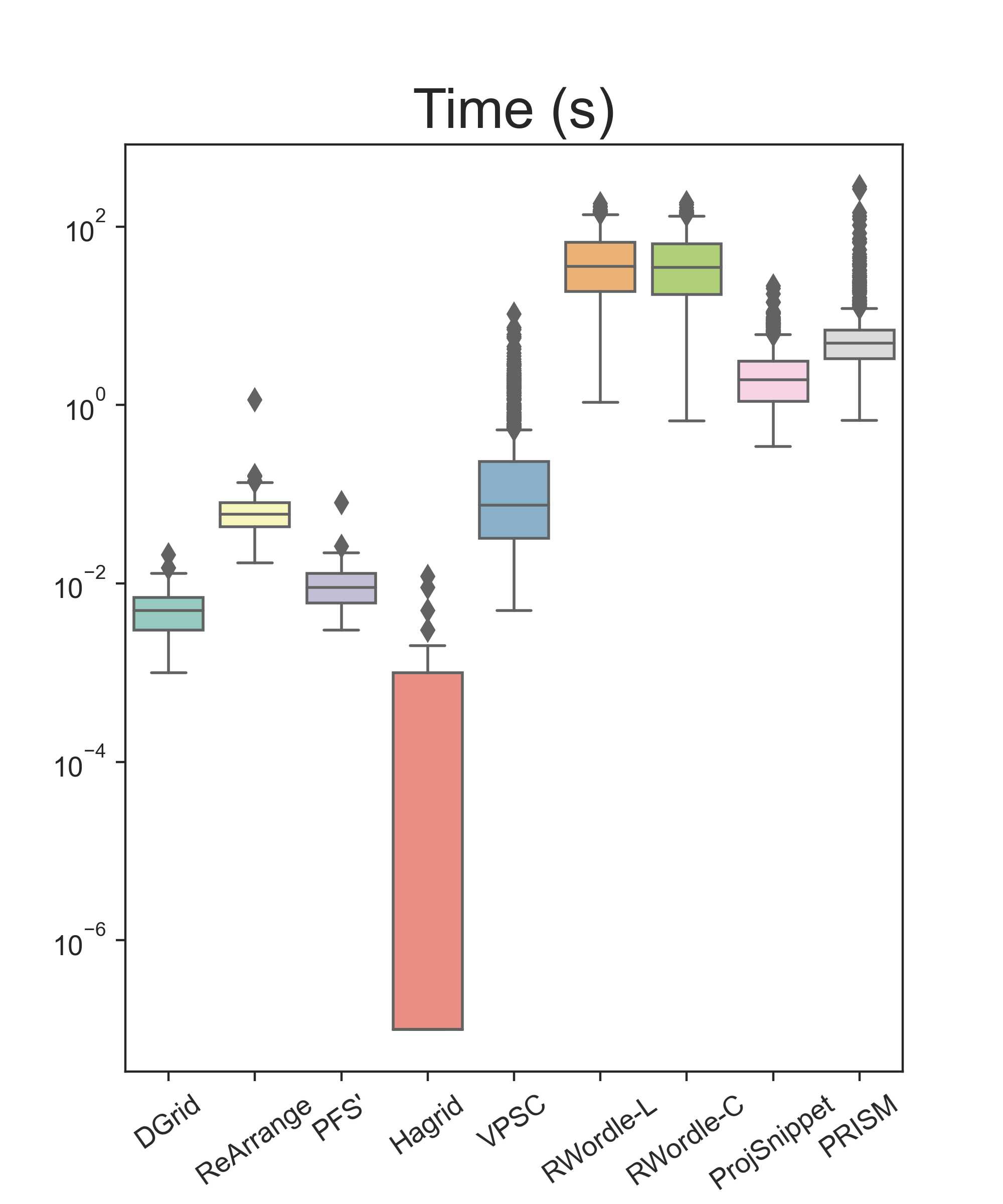}
    \includegraphics[width=.3\linewidth]{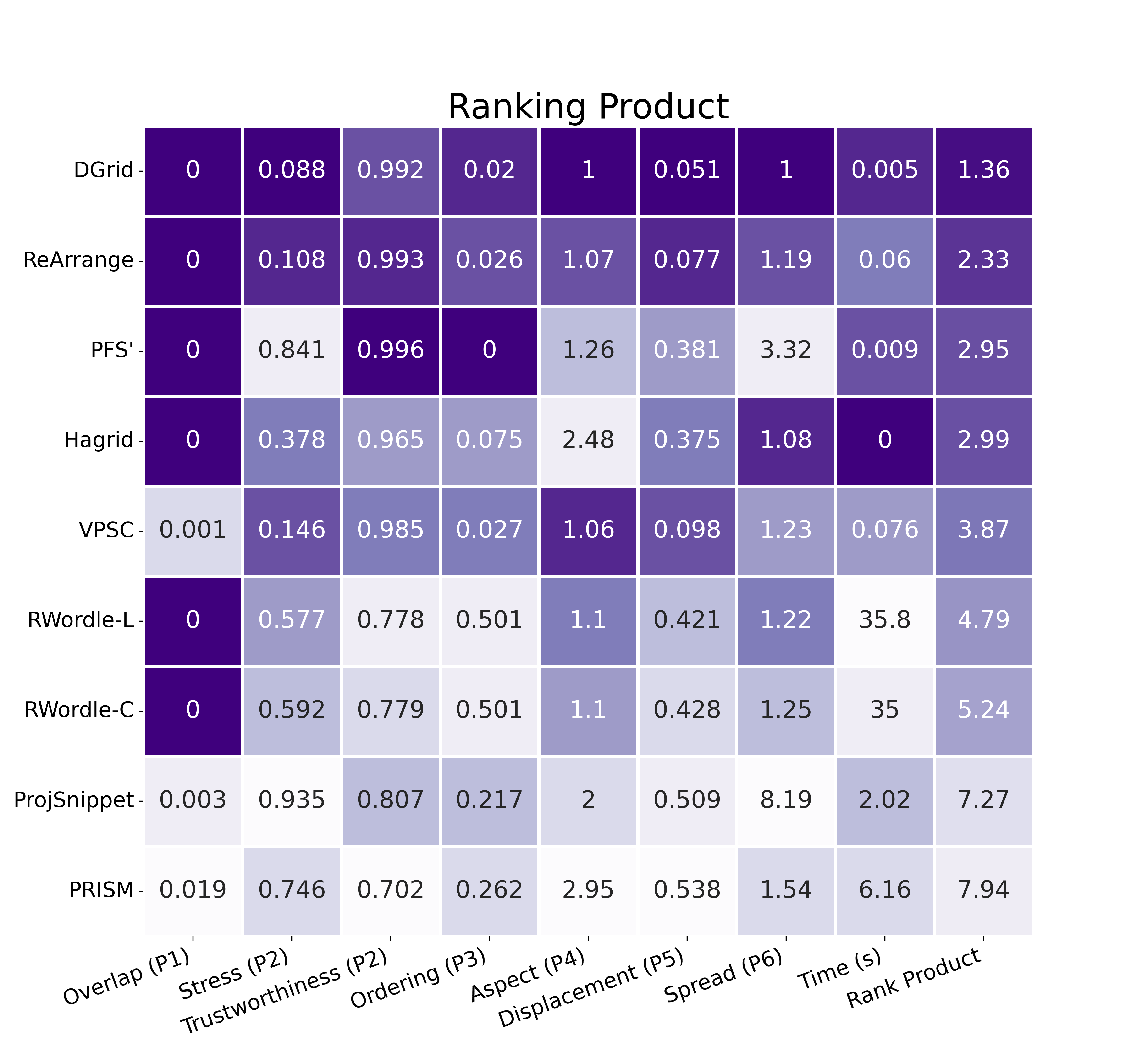}
    \caption{Boxplots summarizing the results of each technique considering the different metrics defined in Sec.~\ref{sec:problem} and the running time. A Ranking Product matrix is also presented, aggregating all the results to rank the techniques according to their median results. The cells' values are the median; the darker the cell, the better the ranking position. Considering the combination of all these metrics, DGrid is the top-ranked technique.} 
    \label{fig:boxplots}
\end{figure*}

\begin{figure}[ht]
    \centering
    \includegraphics[width=.48\linewidth]{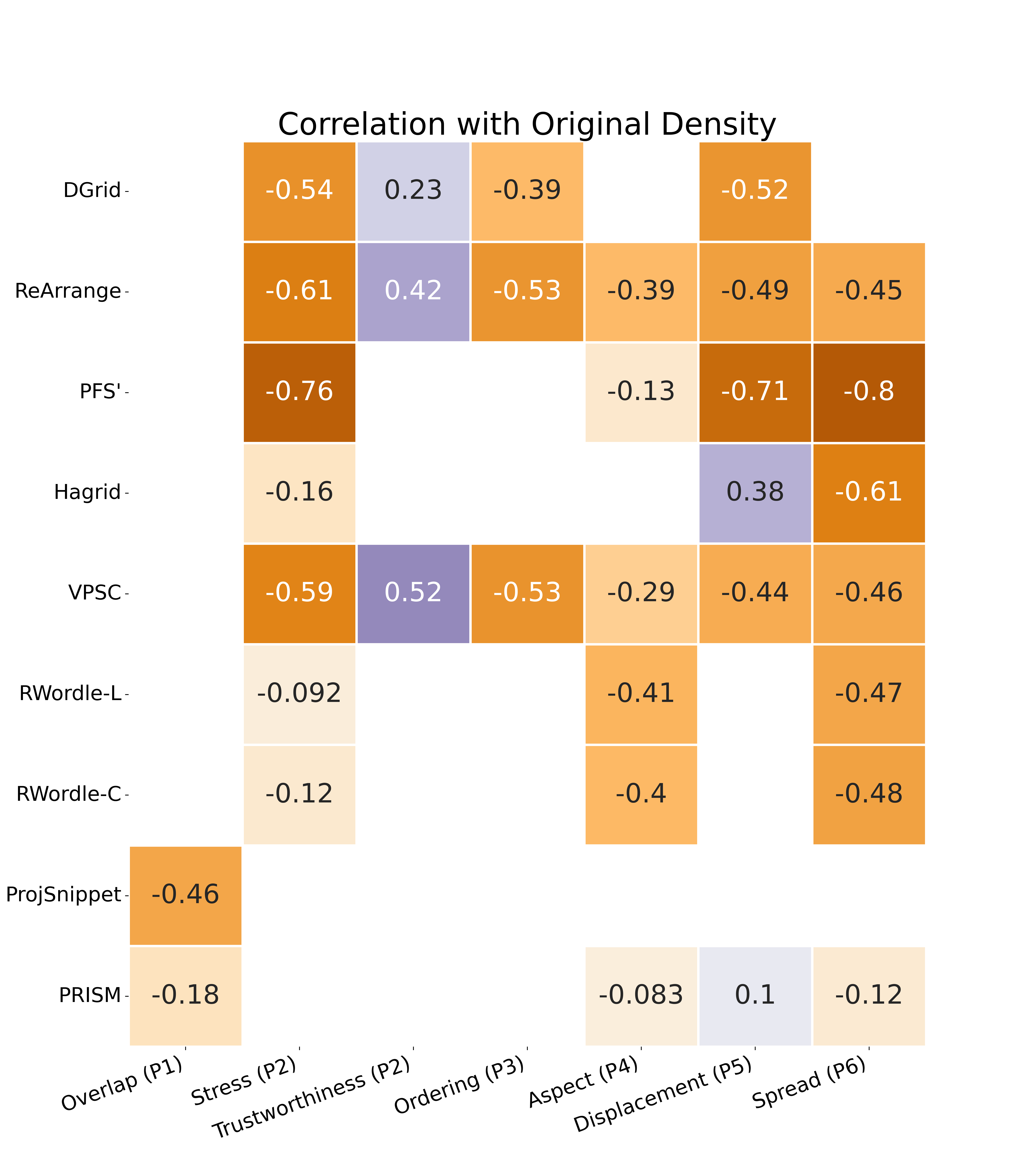}\quad
    \includegraphics[width=.48\linewidth]{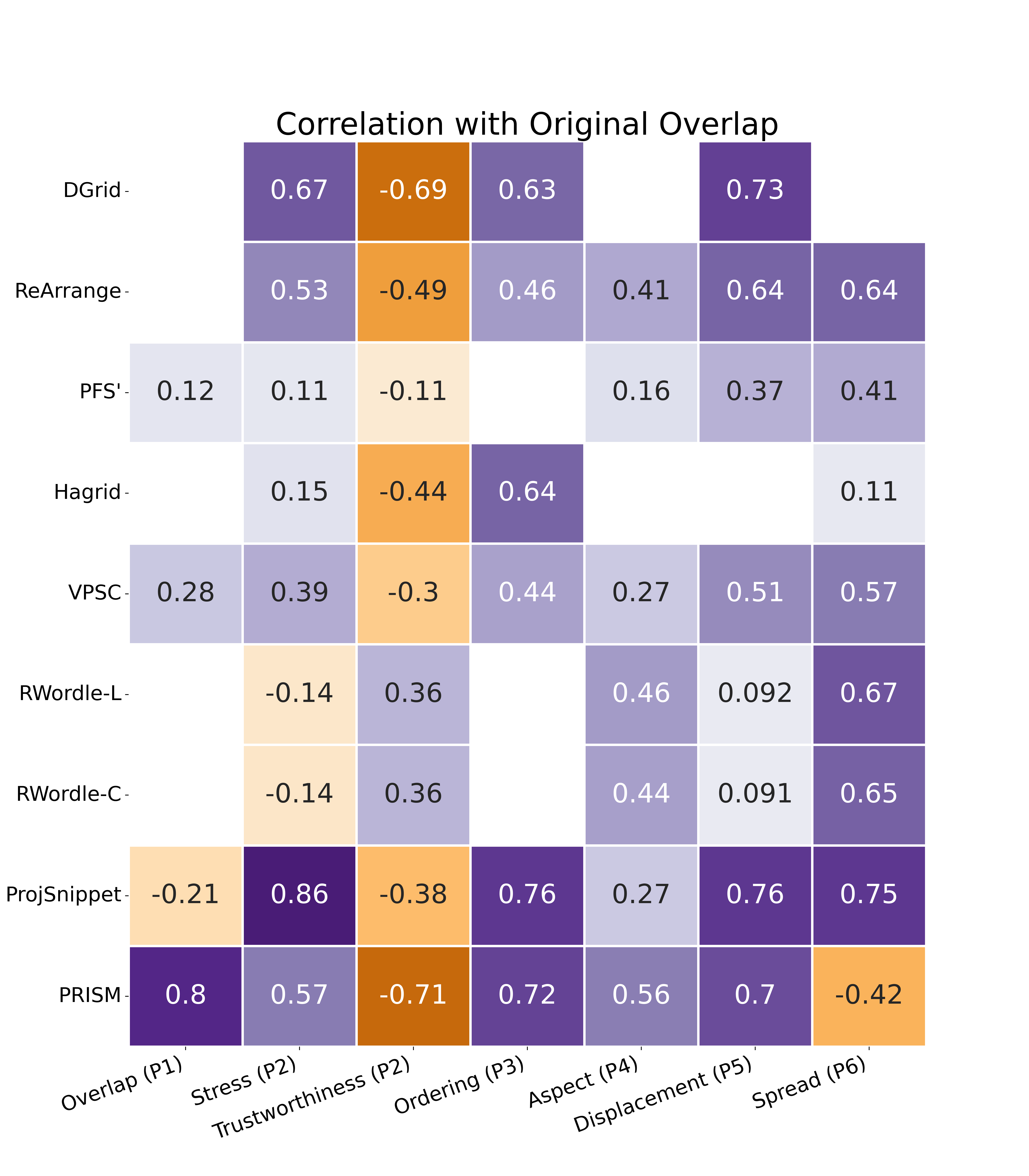}
    \caption{Correlation heatmaps showing how the densities and overlap coefficients of original scatterplots may influence, per technique, the quality of the produced layouts. The missing cells are correlations without statistical significance.}
    \label{fig:heatmap}
\end{figure}

Our analysis is presented in Figs.~\ref{fig:boxplots}~and~\ref{fig:heatmap}. Fig.~\ref{fig:boxplots} presents boxplots summarizing the results of each technique considering the different metrics. Fig.~\ref{fig:heatmap} presents heatmaps with correlations between these metrics and the original scatterplots' densities (Eq.~\ref{eq:mask}) and overlap degrees (Eq.~\ref{eq:overlap}). We use this correlation analysis to understand, per technique, how the different characteristics of the original scatterplots may affect the quality of the produced overlap-free layouts. In the heatmaps, we only show cells with statistical significance ($\rho < 0.01$). Notice that, for these tests, since it is necessary to set the curve level ($l$) for Hagrid, we run it multiple times, varying $l$. We select the value that, on average, minimizes spread so the glyphs, after removing the overlap, maintain (approximately) the same sizes as the original glyphs. In our tests, $l=4.5$.

Interestingly, PFS', Hagrid, VPSC, ProjSnippet, and PRISM were not able to entirely remove overlaps (Fig.~\ref{fig:boxplots}). The problems of ProjSnippet and PRISM have roots in the employed optimization process and numerical stability. We tried to use gradient clipping to alleviate issues with gradient exploding but with limited success. PRISM fails to entirely remove overlaps in $99.5\%$ and ProjSnippet $98.8\%$ of the tested scatterplots, but the overlap removal is nevertheless still high: on average, PRISM removes $80.1\%$ of the overlap in any given scatterplot, and ProjSnippet removes $93.8\%$ of it. PFS' fails to remove overlap in $58.7\%$ and VPSC in $93.4\%$ of the tested scatterplots, but they also still remove most of it: PFS' removes $99.98\%$ and VPSC $98.78\%$ of the overlaps on average for any given scatterplot. We cannot fully understand the causes for PFS' and VPSC techniques to fail. Still, we know that the resulting overlap is positively correlated with the original scatterplots' overlap (Fig.~\ref{fig:heatmap}). As the original overlap increases, the final overlap also increases. Notice that this is also true for the ProjSnippet and PRISM, with a fascinating negative correlation for ProjSnippet (no clues why). For both techniques, the negative result in overlap removal is also correlated with the original scatterplots' density, meaning that these techniques do not work properly when the amount of visual area that is needed to remove the overlaps approaches the total scatterplot area. For Hagrid, we cannot explain why it fails to remove overlaps completely, and no correlation is observed. Nevertheless, it rarely fails, only on $0.8\%$ of the cases, and even when failing, it removes $94.15\%$ of the existing overlap on average. DGrid, Rearrange, RWordle-L, and RWordle-C, completely removed the overlaps in all scenarios. Notice that techniques that do not entirely remove overlaps may have an unintended advantage in a quantitative evaluation considering the metrics of Sec.~\ref{sec:problem}. If a technique does not change the input layout, then stress, trustworthiness, ordering, aspect, displacement, and spread metrics will reach their best values. Therefore, although we ignore this when calculating the other metrics, we advise the reader to keep that in mind.

Regarding aspect ratio preservation (Fig.~\ref{fig:boxplots}), DGrid is the only technique to reach a nearly optimum result. It is the only technique that can preserve original scatterplots' global shapes while Hagrid, PFS', ProjSnippet, and PRISM present the most significant deformations on average. Checking the heatmaps (Fig.~\ref{fig:heatmap}), it is possible to see that DGrid and Hagrid are the only techniques not presenting a statistically significant correlation between original scatterplot overlap and density and the aspect ratio preservation of the produced overlap-free layouts. The others present a positive correlation, indicating that overlap and density may negatively affect the aspect ratio, resulting in distorted layouts. Similar results are presented for spread. DGrid offers a nearly perfect spread, meaning it well preserves original glyph sizes. PFS' and ProjSnippet present the most significant spread, and PRISM often reduces the scatterplot area's size. For most techniques, the spread is correlated to the original density and overlap. Hagrid is the only one in which spread is not correlated with original overlap, indicating that results are not influenced by existing overlap but more by the original density. Since there is no variance in DGrid's results, the spread is not correlated with the original density or overlap. This is not surprising since we intentionally preserve the original aspect ratio and control the displacement by defining our grid dimensions according to the original scatterplot and using it as a constraint to moving points. The other techniques do not set these as hard restrictions, probably not to affect other metrics. Notice that it is possible to relax DGrid's spread by setting $\Delta > 1$, usually improving stress, trustworthiness, and ordering results but affecting displacement. Even with gradient clipping, PRISM could not create bounded overlap-free scatterplots for two scenarios. So we removed them from our analysis. 

Displacement and stress present similar results. DGrid renders the best result, closely followed by ReArrange and VPSC. The others are on the same level of low distance preservation and high displacement. Therefore, DGrid, ReArrange, and VPSC maintain the original scatterplot's general appearance, with low displacement and good global preservation of original distances. Locally, DGrid, ReArrange, PFS', and VPSC present virtually the same results regarding trustworthiness, with PFS' attaining the best result. Not only do these techniques present the best results, but also the deviation is minimal. This indicates that neighborhoods are reliable, a fundamental characteristic since several analytical tasks executed using DR scatterplots are based on neighborhood analysis. As discussed before, displacement, stress, and trustworthiness may be boosted by not removing the overlaps, so VPSC and PFS' may have some advantages for these metrics. However, that effect should be minimal since more than $99.7\%$ of the original overlap is removed. Notice that it is necessary to establish the neighborhood size to calculate trustworthiness. Here we use the common heuristic of setting it to $5\%$ of the scatterplot size.

For the last metric, orthogonal ordering, PFS' renders the perfect result, with DGrid, ReArrange, and VPSC close by, and Hagrid presenting an intermediate result. These techniques have the lowest impact on the user's mental map, meaning that if the user puts side-by-side the original and the transformed layout and focuses the analysis on one point, what is at the top/bottom, left/right of that point, is preserved. However, for most techniques, there is a cost, which is an increased spread.  ReArrange and VPSC increase the scatterplot sizes by $20\%$ while the PFS' increases more than $330\%$. DGrid ensures good orthogonal ordering without incurring expanding the scatterplot area, preserving the perceived glyphs' sizes.  

\begin{figure}
    \centering
    \includegraphics[width=.4\linewidth]{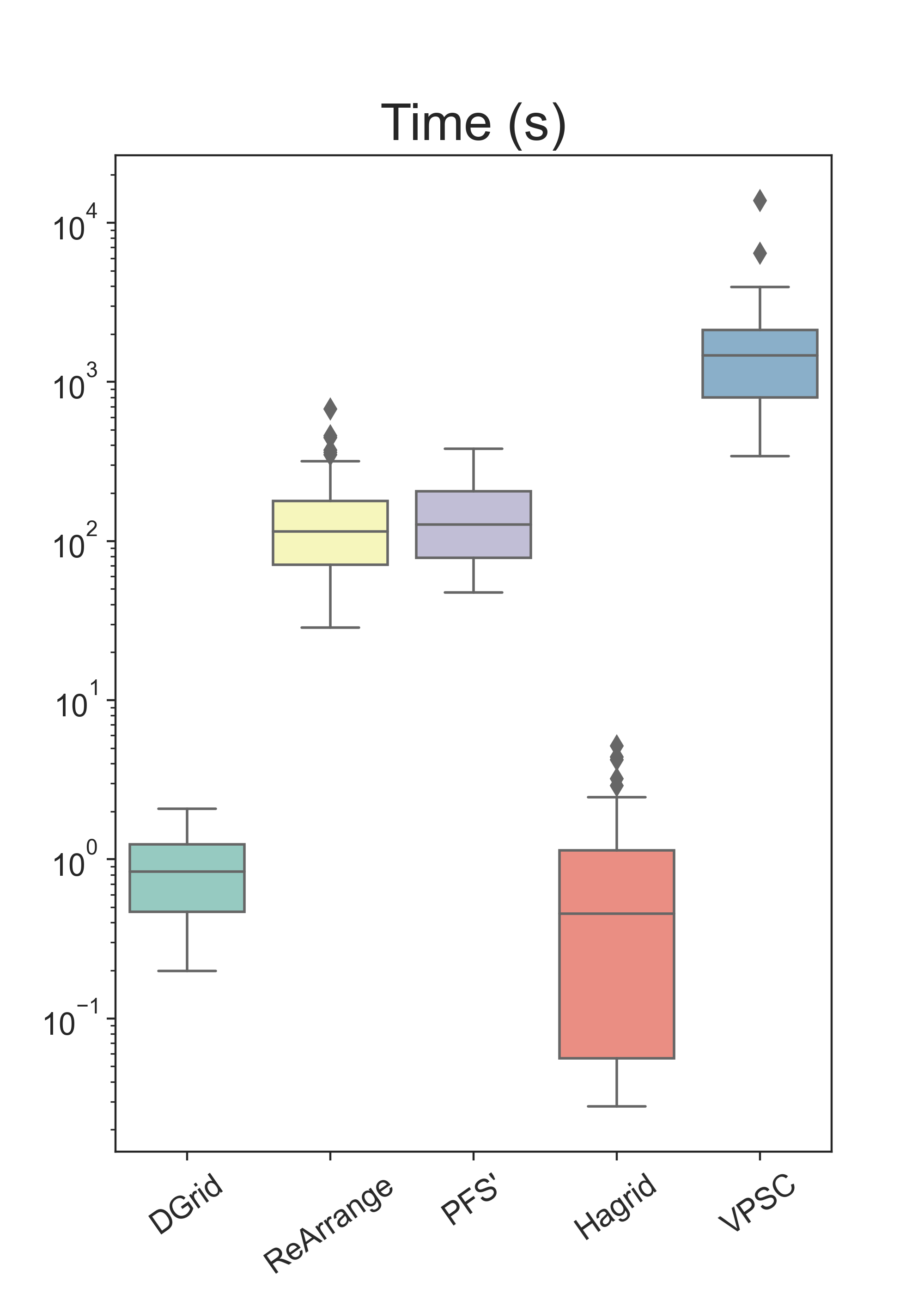}
    \caption{Boxplot summarizing the results considering larger datasets ($50k\sim 100K$ instances). Hagrid is the fastest technique, twice as fast as DGrid. DGrid is two or three orders of magnitude faster than the others.}
    \label{fig:time}
\end{figure}

In terms of running times, even though the datasets are small, two groups of techniques can be observed: one containing techniques that run under a second; and another composed of methods that need more (for some, much more) than that.  Hagrid is the fastest technique, followed by DGrid. ReArrange, PFS', and VPSC practically present the same execution time (the difference is in milliseconds). To show the real differences between these five faster techniques, we create $120$ datasets following the same procedure explained before, but with sizes varying between $50,000$ and $100,000$. Fig.~\ref{fig:time} presents the boxplot summarizing the results. Hagrid (median=$0.46$s) is almost twice as fast as DGrid (median=$0.83$s), and DGrid is at least two orders of magnitude faster than ReArrange (median=$114.89$s) and PFS' (median=$126.91$s) and three orders faster than VPSC (median=$1,471.35$s).

%
% time (s)
% DGrid           0.8340
% ReArrange     114.8925
% PFS'          126.9060
% Hagrid          0.4545
% VPSC         1471.3535

To consolidate these results, we calculate the rank product~\cite{rankproduct2004} using all metrics and running time (for the small datasets). Rank product is a non-parametric statistical method used in biology to combine different ranks and define a general rank. If $r_{t,i}$ denotes the rank position of technique $t$ in the i$^{th}$ metric, the final rank of $t$ considering $K$ different metrics is calculates as $(\prod_i^K r_{t,i})^{1/K}$. For each metric, we create a rank based on the techniques' median values (rounding to the 3rd decimal place). Also, to reduce the advantage of the techniques that do not entirely remove overlaps, we added a penalty to the metrics that is proportional to the amount of overlap that is not removed compared to the original layout. For the minimization metrics (the lower, the better), we use $M_p = (1 + Overlap(\mathcal{P'})/Overlap(\mathcal{P})) \times M$; otherwise, we use $M_p = (1 - Overlap(\mathcal{P'})/Overlap(\mathcal{P})) \times M$, where $Overlap(.)$ is defined in Eq.~(\ref{eq:overlap}), $M$ denotes any metric and the original layout ($\mathcal{P}$) always presents some level of overlap. Fig.~\ref{fig:boxplots} presents a heatmap showing the median results of each technique for each metric, with the cells colored according to the rank position per metric. The darker the color, the best the rank position. DGrid presents the best trade-off among all the techniques and is ranked first in terms of attending to the principles presented in Sec.~\ref{sec:problem}. ReArrange is in second place, followed by PFS' and Hagrid in the third and fourth places (with a minimal difference). VPSC appears in the fifth position, RWordle-L in the sixth, RWordle-C in the seventh, Projsnippet in the eighth, and PRISM in the last (without penalization, PRISM and ProjSnippet would switch places).

Finally, besides testing with the produced artificial layouts, we also quantitatively compared the techniques with scatterplots generated by DR techniques. In this test, we use the scatterplots provided on the survey by Espadoto et al.~\cite{8851280}. The results are detailed in the accompanying supplemental material. In summary, considering the techniques we were able to test (in some cases the scatterplots were too large), DGrid had the best results in most metrics on average, only PFS' is better on ordering and Hagrid on running time. In these two cases, DGrid attained the second-best result. In terms of ranking, the techniques we tested follow the same order of the ranking presented in Fig.~\ref{fig:boxplots}, also attesting to the top-quality of DGrid in a ``real'' scenario, where scatterplots are generated using DR methods.

\subsection{User Evaluation}

Some characteristics of DGrid, such as the produced layouts' general appearance, are inherently subjective and much harder to capture with objective metrics. For example, the impact of aligning the points into a grid, which interferes with their free distribution but may lead to a more organized look and feel, or how the expansion of highly dense regions hinders user perception of groups and other patterns present on the original scatterplots. To account for these aspects, we executed a user test to evaluate how removing overlap affects user perception of the preservation of patterns and the aesthetics of the produced layouts.

The test consisted of two main parts. In the first (named \emph{Similarity Assessment}), given an original scatterplot (with overlap) and nine possible overlap-free alternatives produced by different techniques, the participant was asked to select a minimum of one and a maximum of three overlap-free layouts that best preserved the general characteristics (patterns) of the original, including general scatterplot aspect (width/height), groups, borders between groups, item positions, and item size (circles). This was repeated for 15 different sets of originals + overlap-free alternatives. In the second part (named \emph{Aesthetic Assessment}), given nine overlap-free alternatives produced by different techniques (without the original), the participant was asked to choose the one that is the most aesthetically pleasing and easy to understand (regarding, e.g., separation and boundaries between groups). This was repeated for $10$ different sets of overlap-free alternatives, again with a diverse range of characteristics. More details about the experiments, including all the figures used, can be found in the supplemental material. At the end, we also asked for optional qualitative feedback regarding the reasons for their choices. Each participant assessed $15$ sets of scatterplots in the first part (\emph{similarity}) and $10$ sets in the second part (\emph{aesthetics}), presented with a diverse range of characteristics such as numbers of groups, levels of separation between groups, and densities. The scatterplots were anonymized (i.e., the participant did not know which technique generated them), and the presentation order of the alternatives was always randomized per set and participant. The test was distributed online via \textit{Google Forms} and was performed by $51$ volunteers from various universities in different countries and continents (more information about the participants can be found in the supplemental material).

\begin{figure}[ht]
    \centering
    \includegraphics[width=.49\columnwidth]{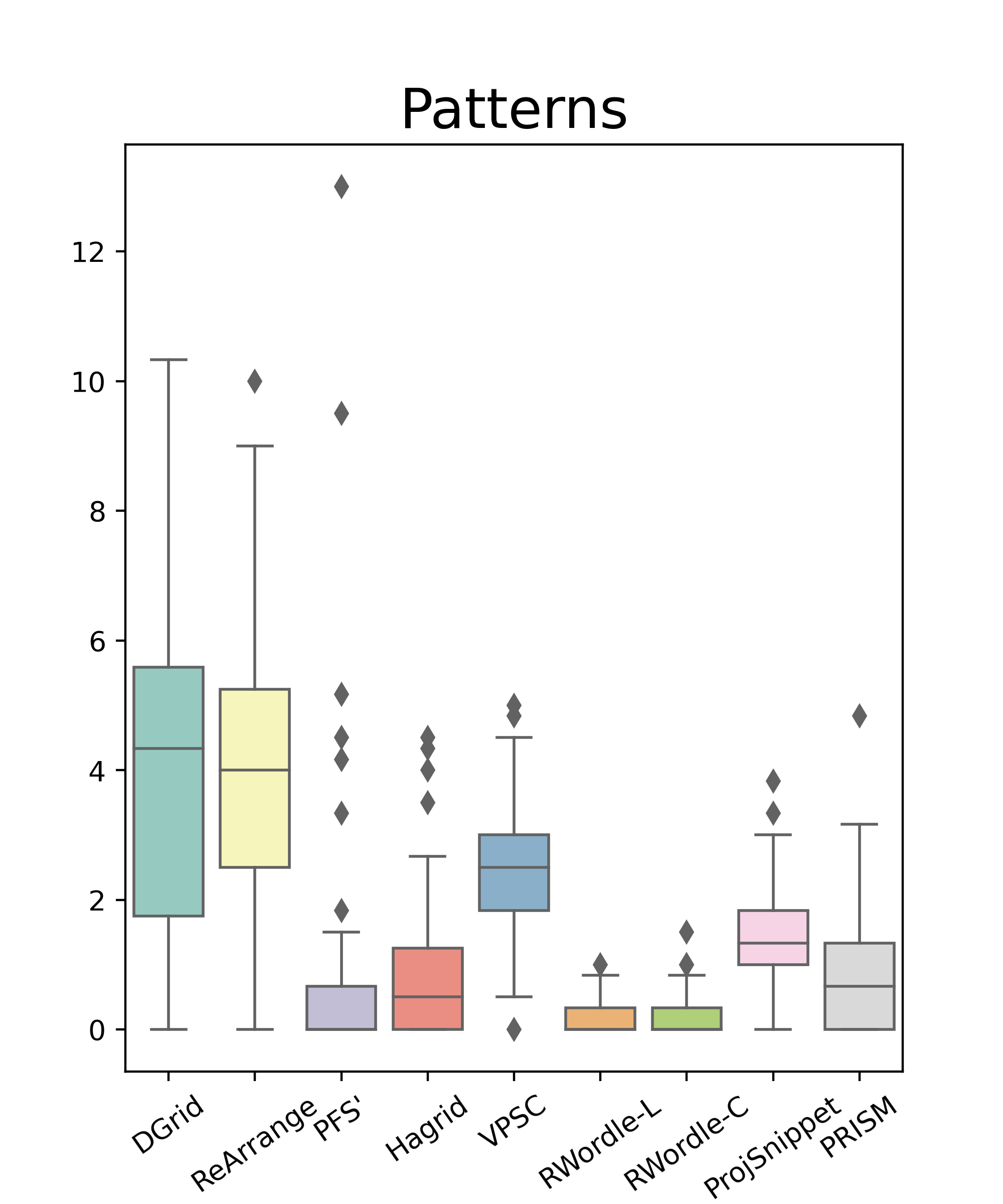}
    \includegraphics[width=.49\columnwidth]{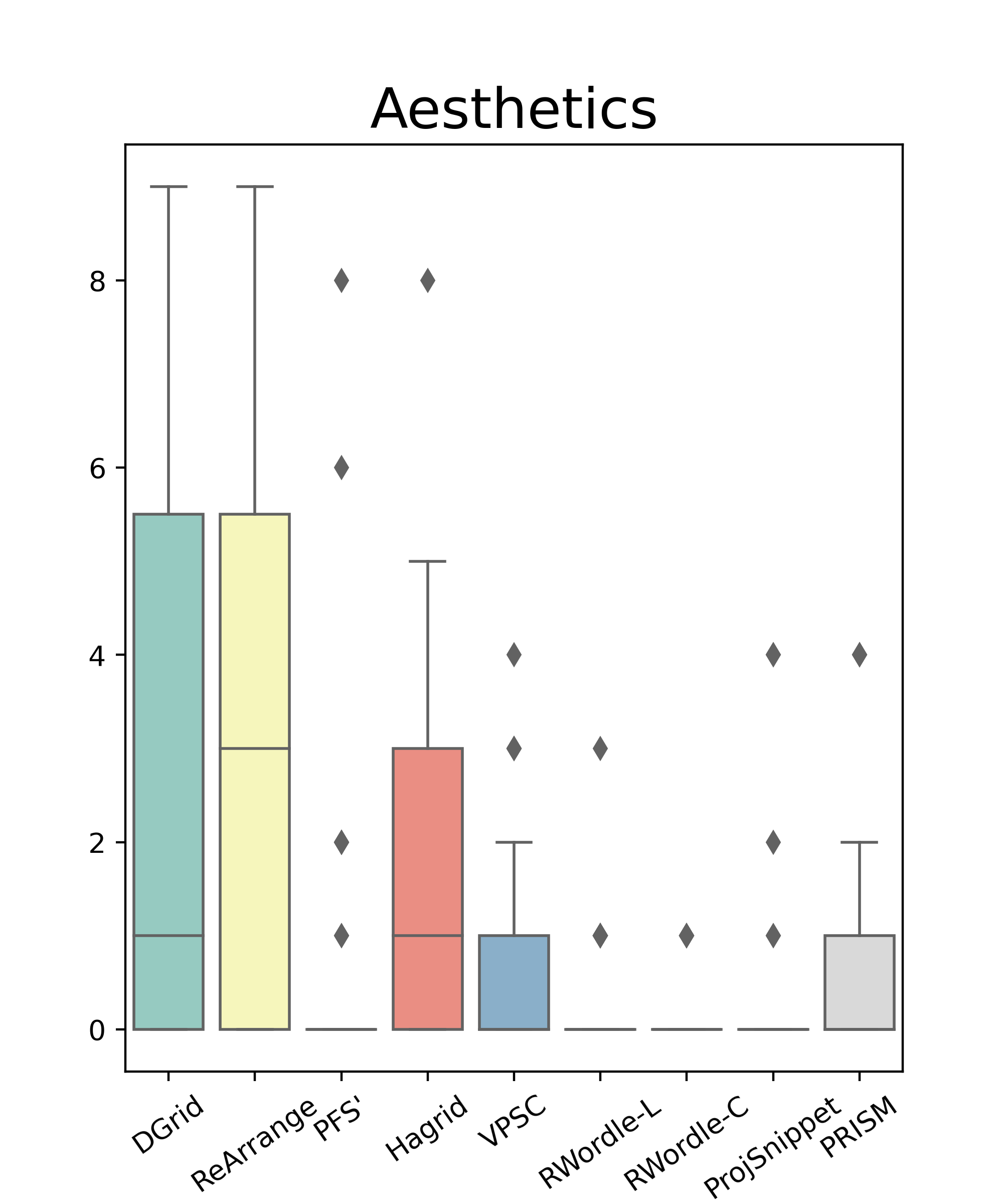}
    \caption{Results of the user evaluation. DGrid and ReArrange have the highest scores for both similarity and aesthetics, with significantly larger scores than all other techniques.}
    \label{fig:user-eval}
\end{figure}

The results---the distributions of scores for each technique in both parts of the user evaluation---are shown in Fig.~\ref{fig:user-eval}. In the first part, we compute a score for each technique (per participant) as follows. Every time a technique is selected as one of the best overlap-free alternatives, it receives a score of $1 / N_{choices}$, where $N_{choices} \in \{1,2,3\}$ is the number of alternatives selected for the same set. Thus, if the participant selected only one alternative, it gets the maximum score of $1$ for that specific set; otherwise, it gets a score of either $\frac{1}{2}$ or $\frac{1}{3}$. The final score of that participant's technique is the sum of the scores for all the $15$ assessed sets of scatterplots. In the second part, each technique's score (per participant) is simply the number of times it was selected as the most aesthetic-pleasing alternative.

DGrid and ReArrange achieve the highest mean scores ($M$) in both parts of the evaluation, with very similar values. For part 1 (\emph{similarity}), ReArrange achieves ($M=3.984, SD=2.126$) and DGrid achieves ($M=3.899, SD=2.655$). Indeed, a one-sided two-sample $t$-test of the null hypothesis that their difference is not statistically significant returns $p=0.429$, failing to support the alternative hypothesis that ReArrange's mean score is larger than DGrid's mean score. On the other hand, when compared to the third largest mean score, VPSC ($M=2.500, SD=1.074$), a similar test returns $p=0.0004$, which indicates that DGrid's mean score is significantly larger than VPSC's. Since the scores dropped sharply further down the ranking, we considered it was not necessary to test further.

For part 2 (\emph{aesthetics}) the conclusions are similar: ReArrange achieves ($M=3.216, SD=3.087$) and DGrid achieves ($M=2.882, SD=3.063$). Again, a one-sided two-sample $t$-test of the same null hypothesis (no statistical difference) returns $p=0.293$, failing to indicate that ReArrange's mean score is significantly larger than DGrid's mean score. This time the third largest mean score belongs to Hagrid ($M=1.353, SD=1.695$), and the one-sided $t$-test between DGrid's and Hagrid's mean score returns $p=0.001$, again indicating that DGrid's mean score is significantly larger than the third place.
 
Finally, we also asked for optional, open-ended qualitative feedback regarding the reasons for participants' choices when selecting layouts considering best similarity preservation and aesthetics. We received 34 responses, and the main themes found in the answers were: (a) clear separation between groups (19/34); (b) simplicity, uniformity, and/or clarity of the layout (8/34); (c) layouts that look more ``natural'' and not too structured (6/34); (d) clear visualization of the density of the clusters (5/34); and (e) as little remaining overlap as possible (3/34). Other themes mentioned were the amount of white space (the less the better) and the avoidance of groups that look too squared.

\section{Discussions and Limitations}
\label{sec:limitations}

\noindent\textbf{Design considerations.} In our solution for assigning points to a grid (Algorithm~\ref{alg:dgrid}), the bisecting process uses a simple heuristic to partition the points, splitting them into sub-partitions with similar sizes. As discussed (Sec.~\ref{sec:assignment}), we split into halves since, without any expensive test, we increase the probability of getting the largest partitions with the desired distribution. We have also tested creating partitions with different sizes, but the results were not statistically different. One alternative to improve could be to use a strategy to find the best (orthogonal) partitioning, for instance, using the Jenks' Natural Break~\cite{Jensk1967IYC}. The Natural Break is a clustering approach that partitions univariate datasets to reduce the variance within-cluster and maximize the variance between clusters. We also tested with that, and the marginal improvement does not justify the increase in complexity and running time. We input two main reasons for this result. First, considering a bisection to produce a (sub)grid with $R$ rows and $C$ columns, we have only $R-1$ possible horizontal cuts and $C-1$ possible vertical cuts. So, there is not much room for defining the best partition. Second, since the ``dummy'' points are distributed following the grid cells, the regions of empty spaces already present perfect uniform marginal distributions, so a different splitting does not change the outcome considerably for scatterplots containing empty spaces. 
    
Another design consideration is why not fix the ``dummy'' points and move only the original points? Although a reasonable idea, fixing ``dummy'' points do not work in practice. Since the process of creating such points does not guarantee that the number of not occupied cells close to a high-density region is equal to the number of original points in that region, the original points can be drastically displaced if the ``dummy'' points do not move. 

\vspace{0.25cm}\noindent\textbf{Space constraints.} Although an effective solution to declutter scatterplot layouts, overlap removal may not be the best option in some cases and for some tasks. Firstly, removing overlaps affects distribution detection tasks. This is true for all techniques that limit the visual area to keep glyphs at a readable size. A simple solution could be to show the original and the overlap-free scatterplots side-by-side. In this case, a precise technique, such as DGrid or ReArrange, is mandatory, so the user's mental map is preserved between visualizations. Secondly, the visual area where the overlap-free scatterplot is rendered needs to allow the removal. It is possible to remove overlaps only if the proportion between the available space and the total glyphs' area is larger than one (see Eq.~\ref{eq:mask}). Indeed, if this proportion is not significant and the original scatterplot presents high-density regions, the produced scatterplot will merge groups and fine-grained groups will be blended, impacting class separation tasks~\cite{7864468}; an issue affecting all overlap removal techniques. One solution is to increase the available space, in our case, increasing $\Delta$, but incurring reduced glyphs' size if the display space is not increased. This is especially problematic for large datasets, where $\Delta$ may need to be too big to accommodate all points and empty spaces. In this situation, a sampling technique~\cite{9226404} may be a better choice to help declutter the visual representation. In fact, the choice between overlap removal or sampling should be based on the analytical task at hand and the data set size. Sampling may be a better choice if the goal is to have an overview of a (large) dataset in terms of group formation. However, if visual space allows and beyond groups, the analytical task also involves inspecting individual instances, overlap removal may be better suitable. One idea to take advantage of both concepts is to use sampling and overlap removal together, for instance, aggregating points before removing the overlaps. This involves several practical challenges, so we leave this idea for future work.

\vspace{0.25cm}\noindent\textbf{Glyph's shape and dimension.} DGrid allocates the same layout area to each glyph (grid cells). Although glyphs' dimensions can vary inside that area, if they are too discrepant, unnecessary space will be added to the overlap-free layout. Similarly, if the shapes of the glyphs deviate too much from the rectangular form (they can indeed assume any form), unnecessary spaces between glyphs may also be created. For circular (like pie-charts), diamond, or hexagon-shaped glyphs, DGrid can be adapted to address this issue by horizontally and vertically translating the grid rows by small fractions of the glyphs' maximum width and height. Although an effective solution, this changes the scatterplot aspect ratio (violating \textbf{P4}), so a better solution would be to design an assignment process that considers different partitions of the space. We leave this as future work. In summary, although the resulting grid alignment showed beneficial, bringing some organization to the visualization, other solutions for removing overlap are preferable if glyphs with significantly varying shapes and dimensions are required. 

\begin{figure}[htb]
    \centering
    \subfigure[Original]{\includegraphics[width=.4\linewidth]{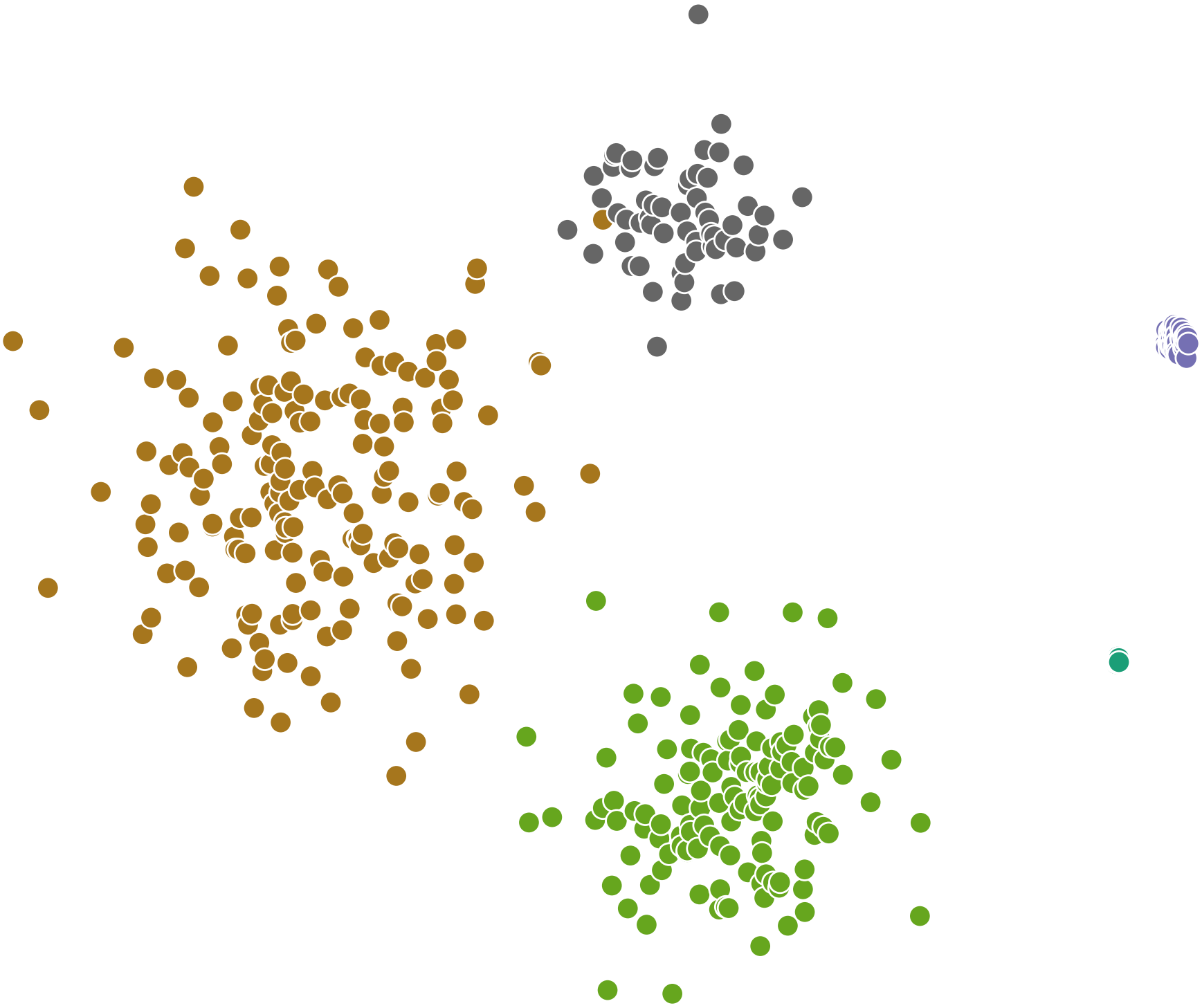}}\qquad\qquad
    \subfigure[$\Delta=1.0$]{\includegraphics[width=.4\linewidth]{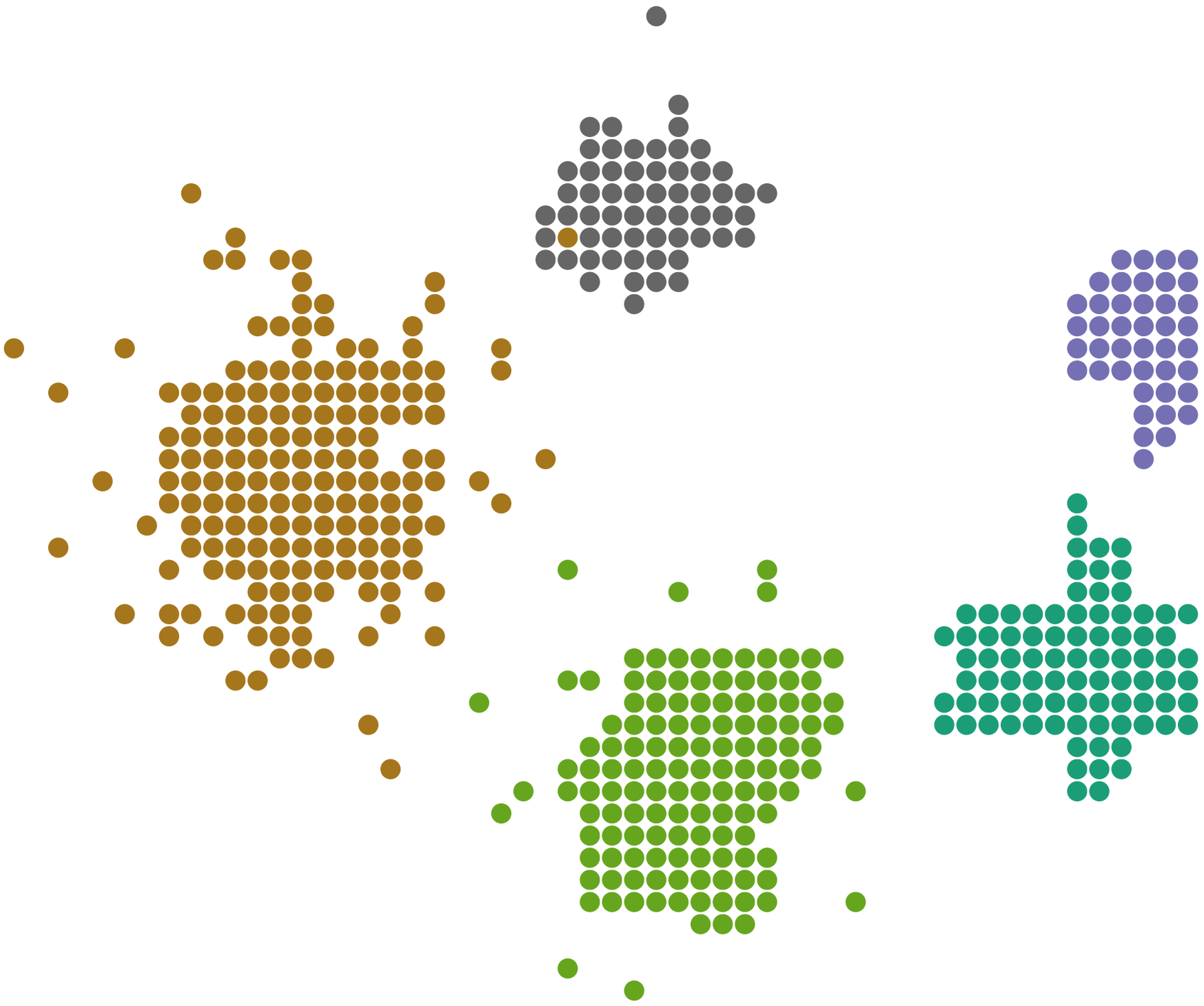}}
    \subfigure[$\Delta=2.0$]{\includegraphics[width=.4\linewidth]{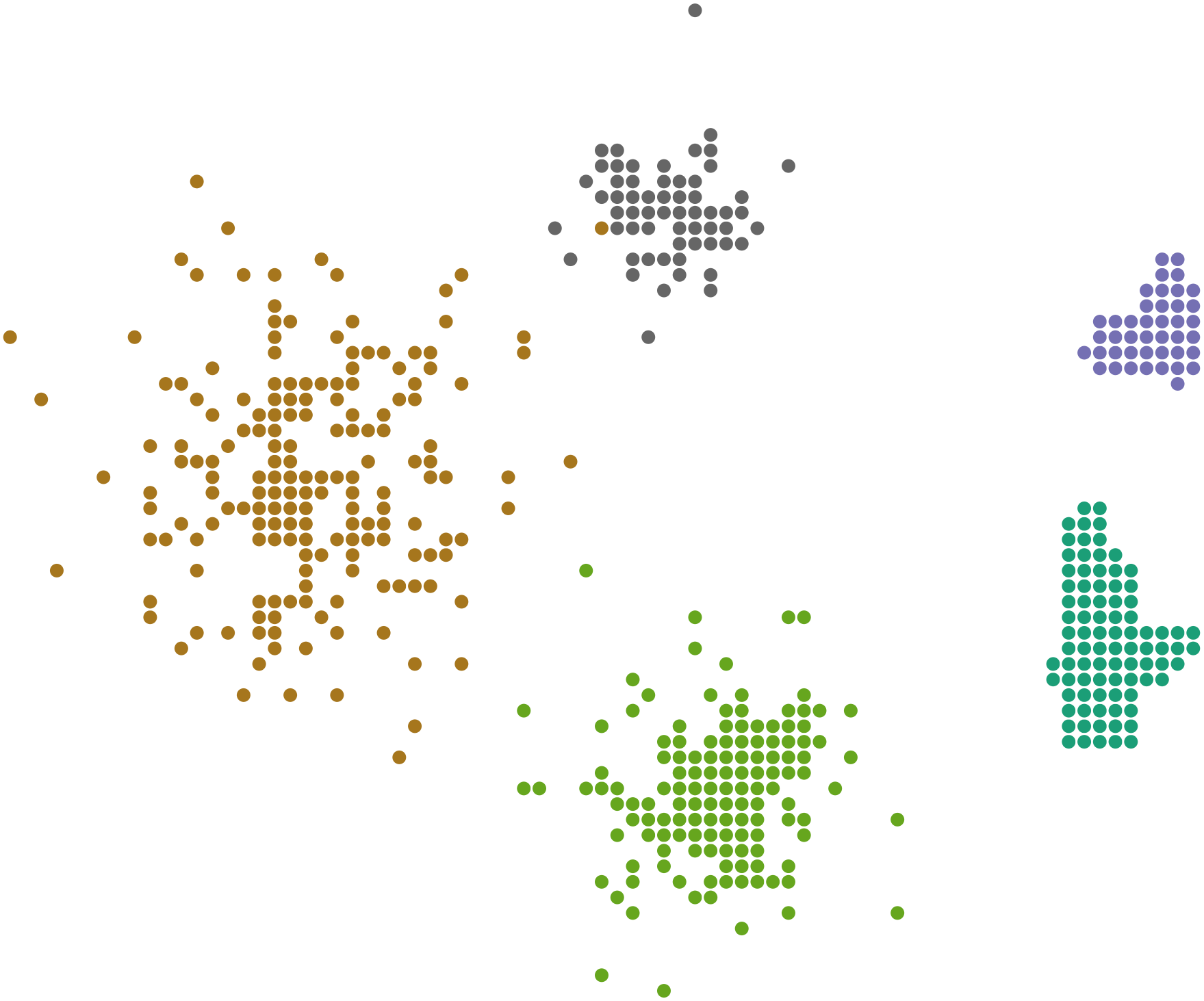}}\qquad\qquad
    \subfigure[$\Delta=4.0$]{\includegraphics[width=.4\linewidth]{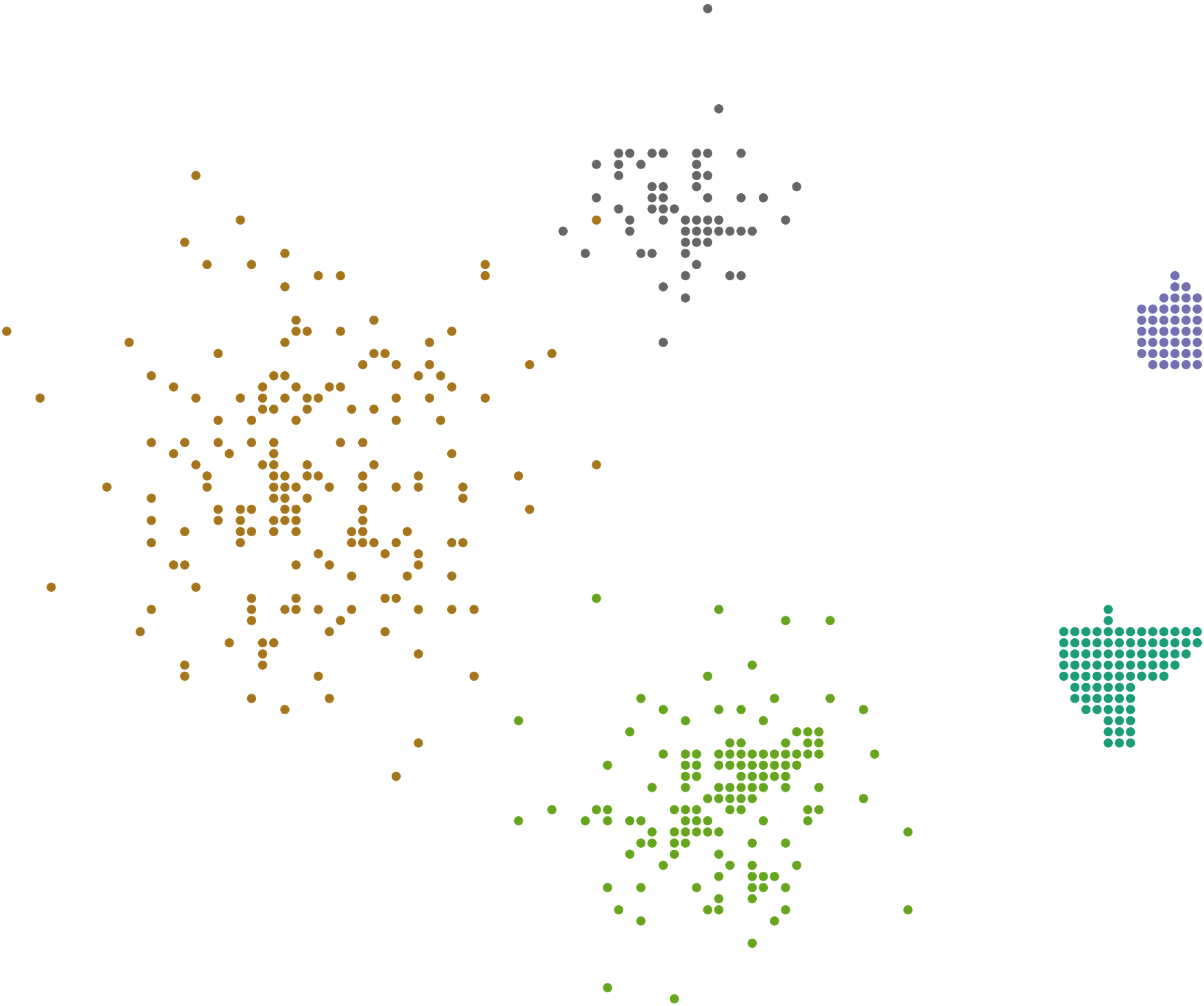}}
    \subfigure[$20\times 25$]{\includegraphics[width=.4\linewidth]{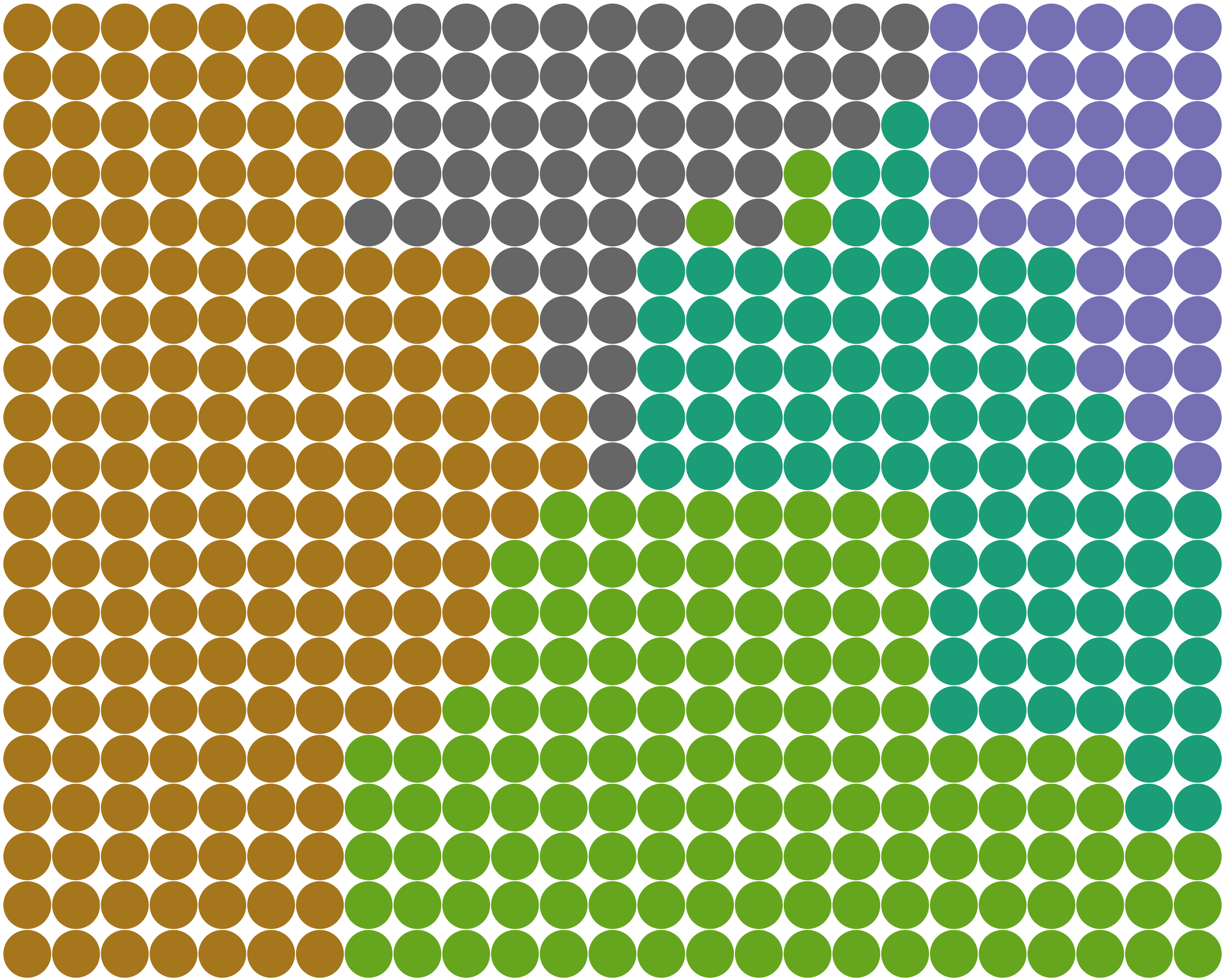}\label{fig:cornercase_manual}}\qquad\qquad
    \subfigure[$\Delta=0.2058$]{\includegraphics[width=.4\linewidth]{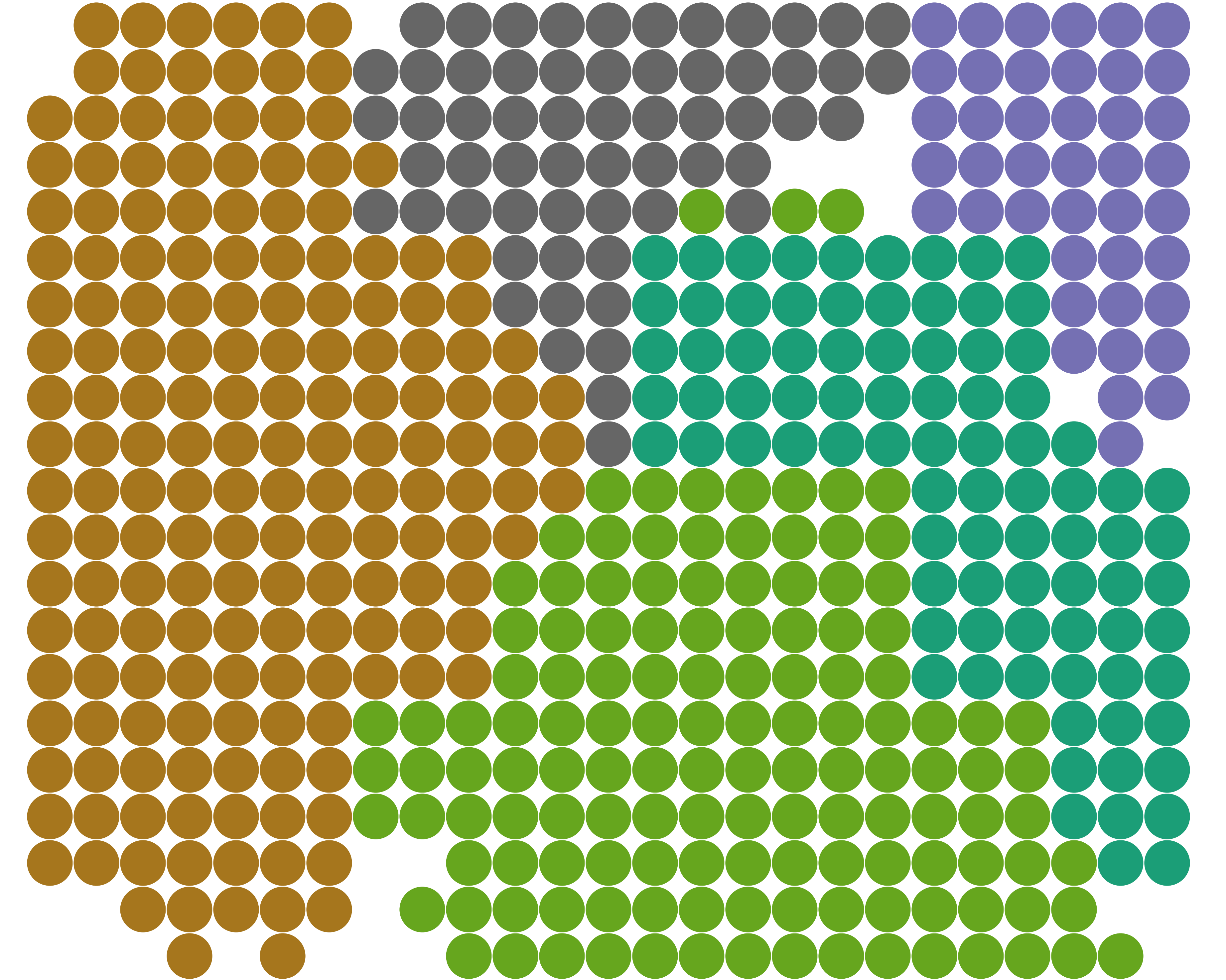}\label{fig:cornercase_delta}}
    \caption{A corner case for removing overlap (a-d). A significant density variation can be observed with two very dense groups (purple and dark green). Although overlap can be fully removed and some distortions can be reduced by increasing $\Delta$, some are still inevitable. Another possibility of using DGrid is to create distance-preserving space-filling grid layouts (e-f), reducing the empty space and increasing glyphs' sizes.}
    \label{fig:cornercase}
\end{figure}

\vspace{0.25cm}\noindent\textbf{Distortions and evaluation.} Overlap removal in scatterplots is a well-documented strategy~\cite{8017602, 6634128}, with different added benefits given the more readable visual representations it generates~\cite{Meuleman2019}. However, it introduces distortions in the produced representations. For example, the scatterplot of Fig.~\ref{fig:cornercase}(a) is an example where it is inevitable to add distortions in the overlap-free layout due to density variation and the presence of very high-dense groups of points (purple and dark green). For DGrid, since we use orthogonal partitioning in the grid assignment process, these high-dense groups, especially those close to the scatterplot borders, tend to have rectangular sides or shapes. This can be mitigated, up to an extent, by increasing $\Delta$ (Fig.~\ref{fig:cornercase}(b-d)) but paying the price of reducing the glyphs and potentially creating unreadable layouts. If it is necessary to detail how distorted the overlap-free layout is when compared to the original layout, different strategies could be used. Some of the metrics presented in Sec.~\ref{sec:problem}, more specifically, stress, trustworthiness, ordering, and displacement, could be mapped, e.g., to color or brightness on top of each scatterplot point, or more visually advanced methods could be used to detail neighborhood and group preservation distortions~\cite{MARTINS201426}.

One trait not explored in this paper is DGrid's ability to decrease (or eliminate) the empty space, producing distance-preserving space-filling grid layouts. This can be done in two different ways. One is completely ignoring the dummy points, using as input only the original scatterplot points and setting the number of rows ($R$) and columns ($C$) of Algorithm~\ref{alg:dgrid} so that the number of grid cells is as close as possible to but larger than or equal to the number of the original scatterplot points ($N$). Fig.~\ref{fig:cornercase_manual} shows the resulting layout for an original scatterplot of $500$ points when $R=20$ and $C=25$. Since $R \times C = N$, no empty space is left, and all available space is filled. Arbitrarily setting the number of rows and columns may result in distorted layouts. To keep as much as possible the original aspect ratio, we could use a different strategy, defining $\Delta= N / (\frac{W_{bb}}{w_{max}} \times \frac{H_{bb}}{h_{max}})$ instead of setting the number of rows and columns explicitly. This formula results in the smallest value for $\Delta$ to accommodate all glyphs while minimizing the empty space. So if there is enough space, $\Delta < 1$, the glyphs increase in size, and the aspect ratio is preserved. In Fig.~\ref{fig:cornercase_delta}, this is exemplified by showing the layout for $\Delta=0.2058$, better preserving the aspect ratio but adding a few dummy points. We did not explore this option throughout the paper since we aim to remove overlaps from scatterplots. However, using a DR layout as input and reducing (or eliminating) the empty space opens the possibility of using DGrid for other scenarios beyond overlap removal.

Finally, although our user test indicates that DGrid, compared to other techniques, presents a good ability to preserve the original patterns (groups, borders between groups, outliers, etc.), the whole concept of overlap removal would benefit from a more extensive evaluation, considering the needs of different users and varied scenarios to verify in the wild the real benefits of an overlap-free layout and what are the introduced caveats. This is a very complex evaluation, beyond our scope, but important future work.

\section{Conclusions}
\label{sec:conclusions}

In this paper, we proposed \textit{Distance-preserving Grid (DGrid)}, a novel approach for overlap removal of Dimensionality Reduction (DR) layouts. DGrid is a two-step approach that completely removes overlap and maintains the original scatterplot aspect ratio and glyphs' sizes while preserving distance and neighborhood relationships, an essential feature for DR layouts. The set of comparisons we provide shows that DGrid outperforms the existing state-of-the-art techniques considering different quality metrics, and is one of the fastest techniques, especially for larger datasets. A user test attests the good results, with participants ranking DGrid layouts among the ones that best preserve original patterns and are aesthetically pleasing. The quality of the produced layouts and the low computational cost render DGrid one of the most attractive methods to date, allowing inter-group and class-outlier analyses that are usually challenging in typical DR layouts.

% if have a single appendix:
%\appendix[Proof of the Zonklar Equations]
% or
%\appendix  % for no appendix heading
% do not use \section anymore after \appendix, only \section*
% is possibly needed

% use appendices with more than one appendix
% then use \section to start each appendix
% you must declare a \section before using any
% \subsection or using \label (\appendices by itself
% starts a section numbered zero.)
%

% \appendices
% \section{Proof of the First Zonklar Equation}
% Appendix one text goes here.
% 
% % you can choose not to have a title for an appendix
% % if you want by leaving the argument blank
% \section{}
% Appendix two text goes here.

% use section* for acknowledgment
\ifCLASSOPTIONcompsoc
  % The Computer Society usually uses the plural form
  \section*{Acknowledgments}
\else
  % regular IEEE prefers the singular form
  \section*{Acknowledgment}
\fi

The authors would like to thank all reviewers who dedicated time to this paper. Your comments and feedback were beneficial and indeed led to substantial improvement in the quality and clarity of this manuscript. We acknowledge the support of the Natural Sciences and Engineering Research Council of Canada (NSERC), CAPES-Brazil, and the Emerging Leaders in the Americas Program (ELAP) with the support of the Government of Canada.

% Can use something like this to put references on a page
% by themselves when using endfloat and the captionsoff option.
\ifCLASSOPTIONcaptionsoff
  \newpage
\fi

% trigger a \newpage just before the given reference
% number - used to balance the columns on the last page
% adjust value as needed - may need to be readjusted if
% the document is modified later
%\IEEEtriggeratref{8}
% The "triggered" command can be changed if desired:
%\IEEEtriggercmd{\enlargethispage{-5in}}

% references section

% can use a bibliography generated by BibTeX as a .bbl file
% BibTeX documentation can be easily obtained at:
% http://mirror.ctan.org/biblio/bibtex/contrib/doc/
% The IEEEtran BibTeX style support page is at:
% http://www.michaelshell.org/tex/ieeetran/bibtex/
%\bibliographystyle{IEEEtran}
% argument is your BibTeX string definitions and bibliography database(s)
%\bibliography{IEEEabrv,../bib/paper}
%
% <OR> manually copy in the resultant .bbl file
% set second argument of \begin to the number of references
% (used to reserve space for the reference number labels box)
% \begin{thebibliography}{1}
% 
% \bibitem{IEEEhowto:kopka}
% H.~Kopka and P.~W. Daly, \emph{A Guide to \LaTeX}, 3rd~ed.\hskip 1em plus
%   0.5em minus 0.4em\relax Harlow, England: Addison-Wesley, 1999.
% 
% \end{thebibliography}

% \bibliographystyle{IEEEtran}
\bibliographystyle{abbrv-doi}
\bibliography{bibliography}

% biography section
% 
% If you have an EPS/PDF photo (graphicx package needed) extra braces are
% needed around the contents of the optional argument to biography to prevent
% the LaTeX parser from getting confused when it sees the complicated
% \includegraphics command within an optional argument. (You could create
% your own custom macro containing the \includegraphics command to make things
% simpler here.)

%\begin{IEEEbiography}[{\includegraphics[width=1in,height=1.25in,clip,keepaspectratio]{mshell}}]{Michael Shell}
% or if you just want to reserve a space for a photo:

% You can push biographies down or up by placing
% a \vfill before or after them. The appropriate
% use of \vfill depends on what kind of text is
% on the last page and whether or not the columns
% are being equalized.

%\vfill

% Can be used to pull up biographies so that the bottom of the last one
% is flush with the other column.
%\enlargethispage{-5in}

% that's all folks
\end{document}